\documentclass[acmtog]{acmart}
\AtBeginDocument{%
  }

\acmJournal{TOG}
\acmVolume{37}
\acmNumber{4}
\acmArticle{111}
\acmMonth{8}

\usepackage{array}
\usepackage{booktabs}
\usepackage{longtable}
\usepackage{pifont}

\begin{document}

\title{A Review of 3D Object Detection with Vision-Language Models}


\author{Ranjan Sapkota}
\authornote{First and corresponding author.}
\affiliation{%
  \institution{Cornell University}
  \city{Ithaca}
  \country{USA}
}
\email{rs2672@cornell.edu}

\author{Konstantinos I. Roumeliotis}
\affiliation{%
  \institution{University of Peloponnese}
  \city{Tripoli}
  \country{Greece}
}

\author{Rahul Harsha Cheppally}
\affiliation{%
  \institution{Kansas State University}
  \city{Manhattan}
  \country{USA}
}

\author{Marco Flores Calero}
\affiliation{%
  \institution{Universidad de las Fuerzas Armadas}
  \city{Sangolqu´ı}
  \country{ Ecuador}
}

\author{Manoj Karkee}
\authornote{Corresponding author.}
\affiliation{%
  \institution{Cornell University}
  \city{Ithaca}
  \country{USA}
}
\email{mk2684@cornell.edu}

\renewcommand{\shortauthors}{Sapkota et al. 2025, A Review of 3D Object Detection with VLMs}

\begin{abstract}
This paper presents a groundbreaking and comprehensive review, the first of its kind, focused on \textbf{3D object detection with Vision-Language Models (VLMs)}, a rapidly advancing frontier in multimodal AI. Using a hybrid search strategy combining academic databases and AI-powered engines, we curated and analyzed over 100 state-of-the-art papers. Our study begins by contextualizing 3D object detection within traditional pipelines, examining methods like PointNet++, PV-RCNN, and VoteNet that utilize point clouds and voxel grids for geometric inference. We then trace the shift toward VLM-driven systems, where models such as CLIP, PaLM-E, and RoboFlamingo-Plus enhance spatial understanding through language-guided reasoning, zero-shot generalization, and instruction-based interaction. We investigate the architectural foundations enabling this transition, including pretraining techniques, spatial alignment modules, and cross-modal fusion strategies. Visualizations and benchmark comparisons reveal VLMs’ unique capabilities in semantic abstraction and open-vocabulary detection, despite trade-offs in speed and annotation cost. Our comparative synthesis highlights key challenges such as spatial misalignment, occlusion sensitivity, and limited real-time viability, alongside emerging solutions like 3D scene graphs, synthetic captioning, and multimodal reinforcement learning. This review not only consolidates the technical landscape of VLM-based 3D detection but also provides a forward-looking roadmap, identifying promising innovations and deployment opportunities. It serves as a foundational reference for researchers seeking to harness the power of language-guided 3D perception in robotics, AR, and embodied AI. A project associated with this review and evaluation has been created at Github Link: https://github.com/r4hul77/Awesome-3D-Detection-Based-on-VLMs
\end{abstract}

\maketitle

\begin{figure}[h!]
  \centering
  \includegraphics[width=0.99\linewidth]{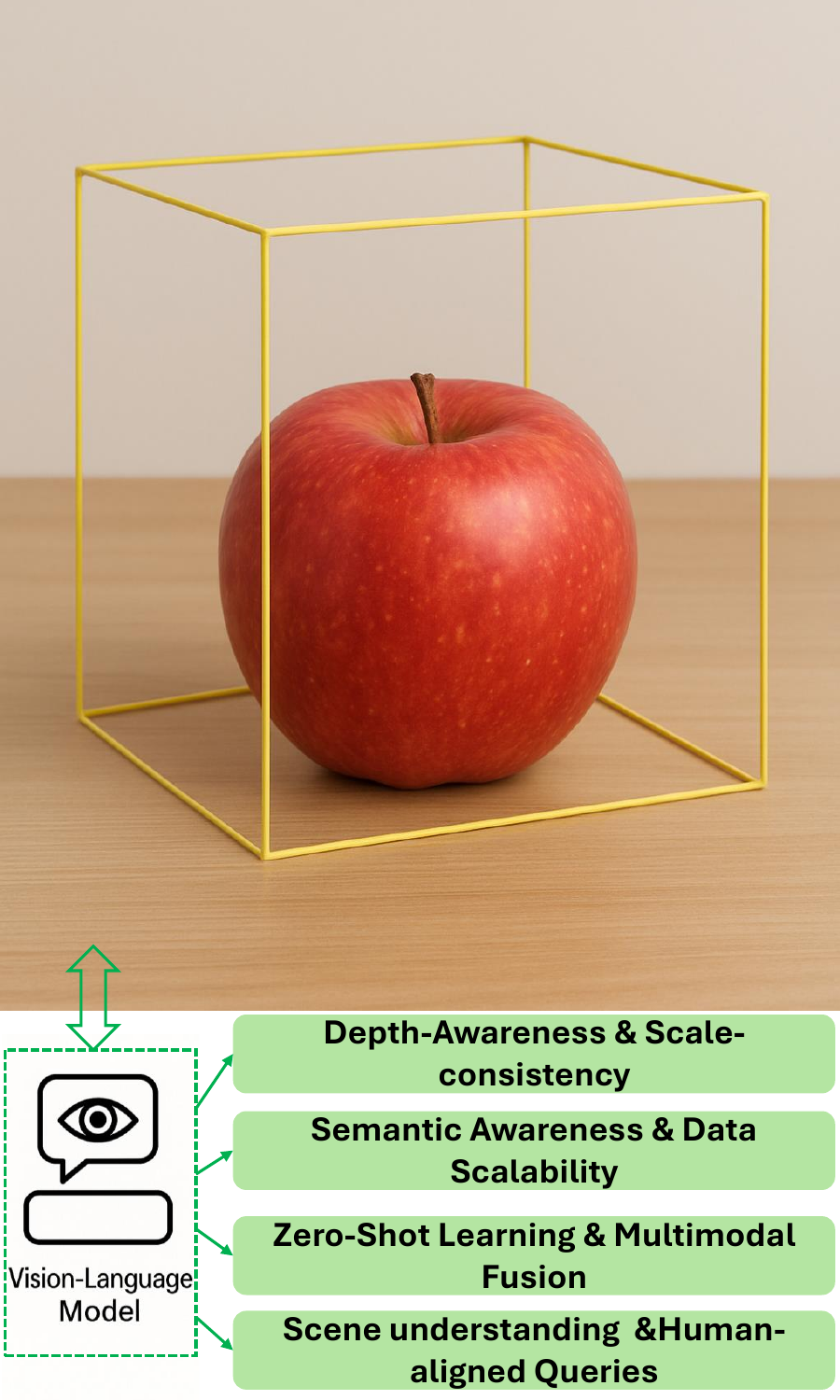}
  \caption{An illustration of VLM-based 3D detection on an apple, showing key advantages including semantic-awareness, zero-shot learning, multimodal fusion, human-aligned queries, and scalable data utilization.}
  \Description{Image1}
  \label{fig:birdview}
\end{figure}
\section{Introduction}
Figure~\ref{fig:birdview} provides a visual example of how object detection has evolved from traditional 2D methods to more advanced 3D detection techniques. The image shows an apple detected in 3D space using a bounding box that captures not just the object’s position on the X and Y axes, but also its depth (Z-axis). This highlights the core benefit of 3D detection: the ability to perceive and localize objects within a spatial environment more accurately than 2D approaches, which are limited to flat image coordinates. As shown in the figure, 2D bounding boxes cannot fully represent an object’s volume, size, or its relation to nearby objects, important factors for tasks such as robotic manipulation or autonomous navigation.

Object detection itself is a foundational task in computer vision, used in a wide range of applications including autonomous vehicles, robotics, surveillance, and augmented reality \cite{ghasemi2022deep}. The primary goal is to identify and localize meaningful objects within images or sensory data by drawing class-labeled boxes around them \cite{zou2023object}. Traditional object detection systems have relied heavily on 2D images, with early models using handcrafted features \cite{papageorgiou1998general, agarwal2002learning}. Modern breakthroughs in deep learning led to the development of powerful real-time detectors like YOLO \cite{redmon2016you}, SSD \cite{liu2016ssd}, and Faster R-CNN \cite{ren2016faster}, which achieve high accuracy even under challenging conditions \cite{sapkota2024yolov10, sapkota2024comparing}. However, without depth information, these 2D systems struggle with tasks that require spatial reasoning or object interaction in a 3D world.

To overcome these spatial limitations, the field has seen a growing shift toward 3D object detection methods that operate on volumetric data such as LiDAR point clouds, depth maps, and RGB-D inputs \cite{wang2021recent}. These models provide a fuller representation of object geometry and spatial relationships, often utilizing voxelization, point-based representations, or 3D convolutions \cite{caglayan2018volumetric, arnold2019survey}. However, while these traditional 3D deep learning models offer significant improvements in spatial reasoning, they come with their own set of constraints \cite{qian20223d, wang2022performance}. They are annotation-heavy \cite{meng2021towards, xiang2014beyond}, requiring large-scale 3D datasets with detailed manual labels that are costly and labor-intensive to create \cite{solund2016large, tremblay2018falling, brazil2023omni3d}. 

Moreover, they frequently suffer from poor generalization across domains \cite{eskandar2024empirical}, lack semantic flexibility \cite{zhang2021objects}, and exhibit rigid retraining requirements \cite{peng2015learning, mao20233d} to adapt to new object categories or deployment settings. Further complicating their deployment, traditional models are often sensor-dependent, relying on carefully calibrated multimodal systems such as LiDAR-camera fusion setups \cite{alaba2022survey}. This makes them brittle in uncontrolled or resource-constrained environments. 

Additionally, these systems are limited in their interpretability and flexibility \cite{wang2022and}, lacking the capability to incorporate high-level, task-oriented instructions \cite{chen20232d}. For example, they cannot interpret human commands like “detect only ripe apples within arm's reach,” nor adapt in real-time to shifting task goals or user intents. The advent of VLMs offers a transformative solution to these challenges. By combining visual perception with natural language understanding, VLMs inject a new level of semantic reasoning into 3D object detection. As demonstrated again in Figure~\ref{fig:birdview}, our illustrated example features a realistic apple being detected in a 3D space, annotated not only with spatial precision but also contextualized through semantic understanding. The figure also highlights the dual sets of benefits: five key advantages of direct 3D detection over 2D methods, and five unique benefits offered by VLM-based 3D detection systems over traditional CNN-based 3D detectors. Direct 3D detection methods outperform their 2D counterparts by offering enhanced spatial accuracy, depth-aware localization, volumetric context, occlusion handling, and metric-scale reasoning. These capabilities are crucial for real-world robotics applications, where a full understanding of object placement and interaction is necessary. At the same time, VLM-based approaches further elevate this by enabling prompt-based control \cite{tang20253d}, zero-shot generalization \cite{zhang2024agent3d}, semantic awareness \cite{zhang2024vla}, multimodal integration \cite{zang2025contextual}, and interpretability \cite{yellinek20253vl, raza2025vldbench}. Unlike CNN-based systems, VLMs can process natural language queries and integrate high-level reasoning into the detection process \cite{chen2024spatialvlm, fu2025scene}, allowing for tasks like “find the nearest apple suitable for picking” without needing specific training for each new task.

Despite the advances in traditional 3D object detection methods, often based on point cloud processing with convolutional neural networks (CNNs), these approaches still face critical limitations in terms of semantic understanding, data efficiency, and adaptability. As highlighted earlier, CNN-based 3D detectors often require expensive sensor calibration, large annotated datasets, and rigid retraining procedures to adapt to new environments. More importantly, they lack the semantic interpretability needed for complex reasoning tasks, such as identifying specific object attributes or responding to user-defined prompts in natural language.

\begin{figure*}[h!]
  \centering
  \includegraphics[width=0.99\linewidth]{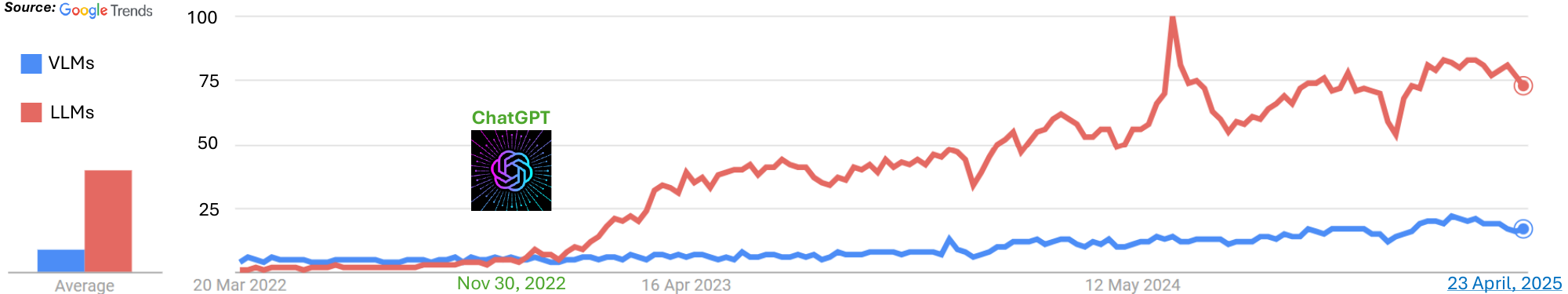}
  \caption{Temporal trend analysis highlighting the surge in global attention toward LLMs and VLMs. Following the public launch of ChatGPT on November 30, 2022, there has been a marked increase in interest and adoption of VLMs, particularly in domains such as 3D object detection, reflecting a shift toward multimodal and prompt-driven AI systems.}
  \Description{Imagetrend}
  \label{fig:trends}
\end{figure*}

In recent years, VLMs have emerged as a transformative solution to the inherent limitations of traditional 2D and geometry-only 3D object detection methods. VLMs integrate the visual pattern recognition strengths of computer vision with the semantic reasoning capabilities of LLMs, allowing for a richer, multimodal understanding of scenes \cite{ma2024llms}. As illustrated in Figure~\ref{fig:trends}, the public launch of ChatGPT on November 30, 2022, sparked an unprecedented surge in interest in LLMs, rapidly establishing them as foundational tools in natural language understanding. While the rise of VLMs has been more gradual, the influence of LLMs has catalyzed increased attention toward multimodal systems. This growing interest is especially evident in domains requiring high-level reasoning and contextual comprehension, such as 3D object detection in robotics, augmented reality, and autonomous navigation. Unlike conventional 3D detectors that rely heavily on point cloud data and annotated datasets, VLMs enable flexible querying, few-shot or zero-shot learning, and semantic task generalization—all without the need for retraining on each new task.

This paradigm shift is exemplified in several recent works. For instance, Agent3D \cite{zhang2024agent3d} uses VLMs to perform open-vocabulary 3D detection based on user-defined queries, enabling robots to locate objects like "the red cup behind the chair" in complex indoor scenes. Similarly, SpatialVLM \cite{chen2024spatialvlm} introduces a spatially-grounded vision-language framework that reasons over RGB-D inputs to detect and describe objects in 3D with spatial context. These models not only identify objects but also reason about spatial relationships, affordances, and human-centric goals. This elevates 3D perception from a purely geometric task to a cognitively informed one, where the model can understand instructions like “find the ripe apple near the bottom-left of the tree.” While VLM trends have not spiked with the same intensity as LLMs, they show a steady and promising upward trajectory, marking the early stages of a broader shift toward truly intelligent, semantically aware 3D systems. As the capabilities of multimodal models expand, VLMs are poised to play a central role in bridging perception and cognition across AI-driven applications.

\textbf{Objective of the Review.}
The objective of this review is to systematically investigate the evolving landscape of 3D object detection using VLMs and to assess their growing role in bridging spatial perception with semantic understanding. We begin by examining the foundational concepts of 3D object detection, its evolution from geometry-based methods to multimodal frameworks, and the distinction from conventional 2D detection. Next, we explore traditional 3D detection architectures—including PointNet++ \cite{sheshappanavar2020novel}, VoxelNet \cite{sindagi2019mvx, chen2023voxelnext}, and PV-RCNN \cite{shi2020pv} to establish a baseline for comparison. Building upon this, we review state-of-the-art VLM-based approaches, highlighting their open-vocabulary capabilities, semantic grounding, and cross-modal alignment. We delve into the underlying architectures, pretraining and fine-tuning strategies, and the visualization of detection outputs to understand how VLMs perceive and reason in 3D space. This review further compares traditional and VLM-based paradigms, identifying the strengths, limitations, and trade-offs of each. Crucially, we analyze current challenges in data availability, grounding accuracy, and computational scalability, while proposing potential solutions such as multimodal dataset expansion and hybrid model integration. By reviewing over 100 papers, this study aims to provide a comprehensive roadmap for researchers, offering insights into the present capabilities and future directions of 3D object detection empowered by VLMs.

\section{Methodology}
\begin{figure*}[h!]
  \centering
  \includegraphics[width=0.90\linewidth]{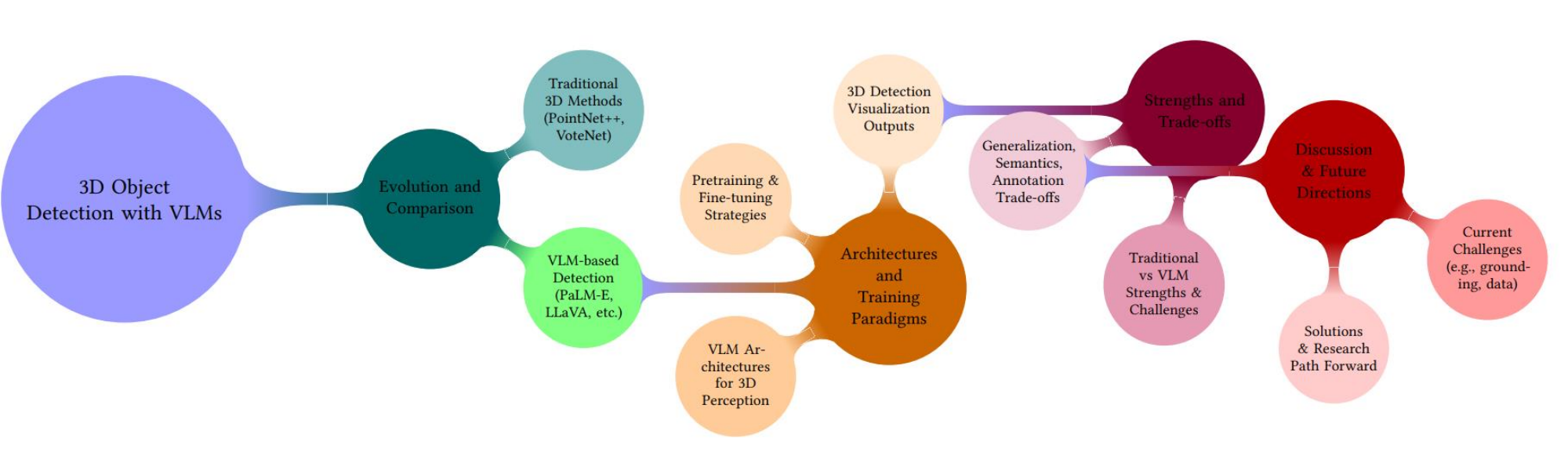}
\caption{Conceptual overview of the methodological and analytical framework used in this review. The mindmap illustrates the study's structure: starting with a comparison of 3D detection methods, followed by VLM-specific architectures, strengths and trade-offs, and ending with a discussion on challenges and future directions.}
\label{fig:mindmap_3dVLM_updated}
  \Description{Imagerg}
\end{figure*}

The methodology adopted for this review is outlined in Figure \ref{fig:mindmap_3dVLM_updated}, which presents our systematic search strategy and paper selection process. This framework ensured a balanced inclusion of studies across traditional 3D object detection, deep learning, and VLM-based approaches. Figure b highlights the rise of research activity in 3D object detection powered by VLMs, peaking in early 2025. Synthesized insights were organized along four critical axes outlined in Figure \ref{fig:mindmap_3dVLM_updated}, tracing the conceptual development and comparative evaluation of 3D object detection paradigms.

\begin{figure*}[h!]
  \centering
  \includegraphics[width=0.83\linewidth]{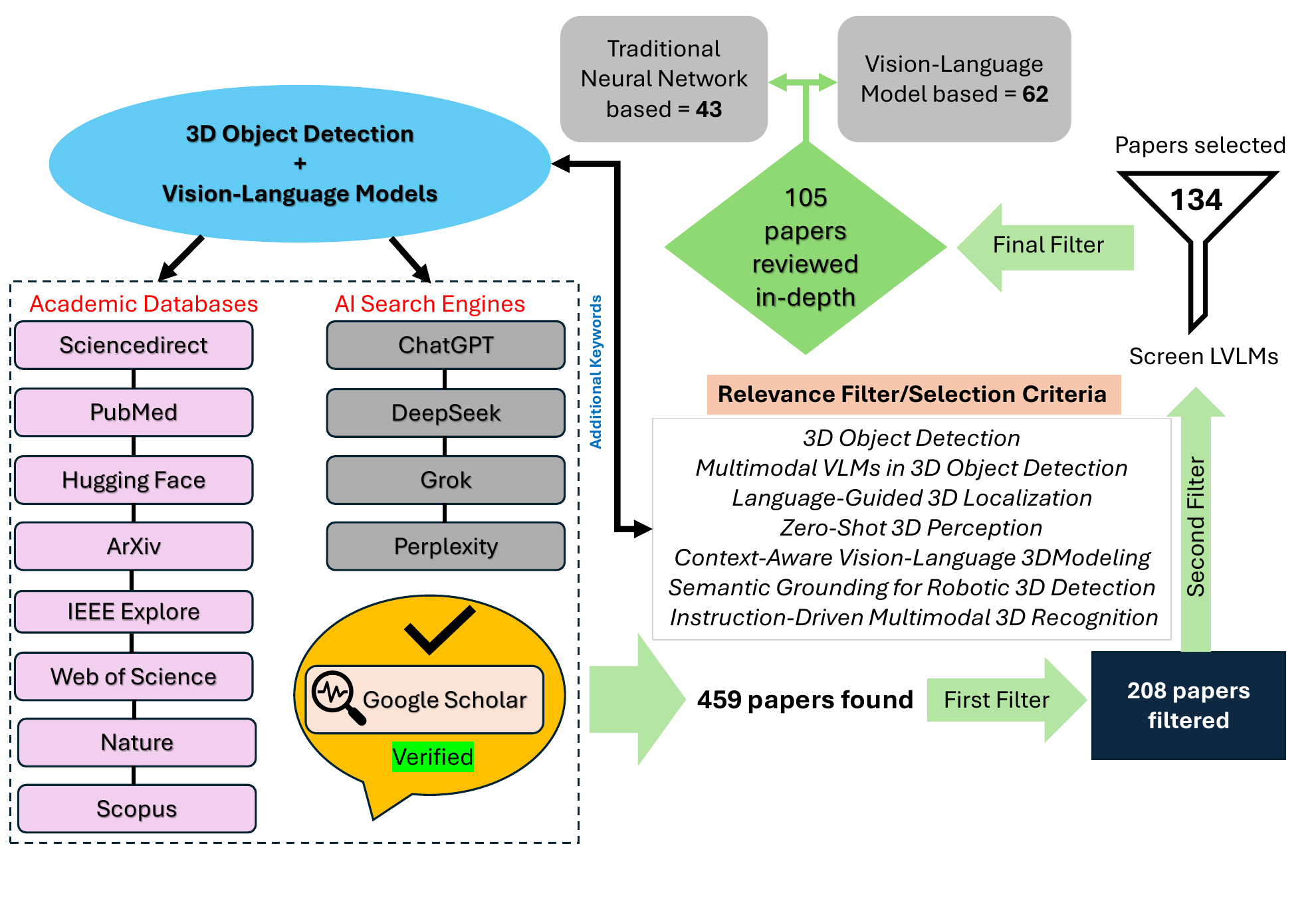}
  \caption{Methodology diagram of this review paper illustrating the hybrid academic and AI-based search strategy, filtering process, and final paper selection, reducing 459 initial papers to 105.}
  \Description{Image2}
  \label{fig:search}
\end{figure*}

\subsection{Search Strategy}
To conduct a comprehensive and systematic review of the 3D object detection landscape, particularly at the intersection with VLMs, we adopted a novel dual-strategy approach that integrates both academic databases and AI-powered search engines—illustrated in Figure~\ref{fig:search}. This marks the first effort in the domain to leverage this hybrid methodology for collecting state-of-the-art literature.
We began by querying twelve prominent platforms, including academic databases such as Google Scholar, IEEE Xplore, Scopus, arXiv, ScienceDirect, PubMed, and Web of Science, along with advanced AI-based engines like Hugging Face, ChatGPT, DeepSeek, Grok, and Perplexity. Our core search terms included combinations and Boolean expressions such as “3D Object Detection,” “Vision Language Models,” “Large Vision Language Models,” “VLMs,” “LLMs,” and “Robotics.” Key queries such as “3D Object Detection + VLMs + Robotics” were designed to capture literature that spans traditional deep learning paradigms and their extension into multimodal perception systems.
The initial search yielded 459 papers. In the first filtering stage, papers were assessed for relevance based on titles and abstracts, narrowing the pool to 208. A second round of filtering, focused on methodological soundness and clarity of application, reduced the count to 134. Finally, a rigorous evaluation of novelty, completeness, and technical relevance led to a curated selection of 105 papers. Among these, 43 focused on traditional neural network-based 3D object detection, while 62 addressed approaches using Vision-Language Models. This refined dataset formed the basis for our structured analysis and review.

\subsection{Literature Evaluation and Inclusion Criteria}

Each selected paper was reviewed for the following criteria: (1) clear methodological contribution to 3D object detection, (2) empirical validation or real-world demonstration, (3) architectural or computational novelty, and (4) relevance to robotics, automation, or agricultural contexts. Specific attention was paid to the subset of papers that used VLMs or multimodal transformers to enhance 3D spatial understanding. We also included several foundational works from 2017–2021 (e.g., PointNet++, VoteNet, SECOND, PV-RCNN, 3DSSD) as the comparative baseline for traditional neural network methods. This helped ground our discussion of the **seven limitations** seen in conventional 3D detection systems—\textbf{Annotation-Heavy, Poor Generalization, No Semantics, Rigid Training, Sensor Dependency, No Prompting}, and \textbf{Limited Flexibility}—relative to more recent VLM-powered approaches.

\subsection{Synthesis and Comparative Analysis}
This review systematically synthesizes and compares the current state of 3D object detection through the lens of VLMs, following a structured flow from literature selection to architectural and application-based evaluations. The analysis begins with the identification of relevant studies using a hybrid academic and AI-based search strategy, as detailed in the methodology. Based on inclusion criteria focused on technical clarity, novelty, and domain relevance, 105 high-quality papers were selected and analyzed. Our results are organized to show the evolution from traditional deep learning approaches to modern VLM-based systems. Traditional methods—such as PointNet++, VoteNet, and PV-RCNN—primarily rely on geometric features derived from point clouds or voxelized data. These methods offer high accuracy in structured environments but face limitations in semantic understanding and generalization. 

In contrast, VLM-based approaches (e.g., DetGPT-3D, ContextDET, and task-conditioned LVLMs) integrate textual prompts with visual-spatial inputs, enabling open-vocabulary detection, instruction following, and zero-shot generalization. These systems show particular promise in robotics applications requiring nuanced reasoning, such as navigation, grasp planning, and task-specific manipulation, especially in unstructured settings. The architecture-focused section highlights foundational components of VLMs for 3D detection, including multimodal fusion layers, pretraining strategies, and fine-tuning mechanisms. Visualization techniques are also discussed to explain how VLMs ground language into spatial contexts. Our comparative analysis evaluates both traditional and VLM-based systems across key criteria—annotation dependence, interpretability, data efficiency, and computational cost. While VLMs excel in semantic grounding and flexibility, they introduce new challenges such as cross-modal alignment, increased computational demands, and sensitivity to noisy inputs. The discussion section outlines these limitations and explores potential solutions such as lightweight VLM architectures, better multimodal grounding techniques, and more robust pretraining datasets.

\section{Results and Analysis}
\subsection{Evolution of 3D Object Detection with VLMs}
\subsubsection{3D Object Detection with Traditional Methods:}
3D object detection emerged as a significant area of research in computer vision with the increasing availability of depth sensors and LiDAR technology in the early 2010s \cite{taylor2008visual, zhang2010multi}. Initially driven by applications in robotics and autonomous vehicles, the field gained traction with the introduction of datasets like KITTI in 2012 \cite{geiger2013vision}, which provided annotated LiDAR and camera data for benchmarking 3D perception tasks \cite{yebes2015visual}. Early methods relied heavily on hand-crafted features and geometric reasoning to infer object locations and shapes in three-dimensional space \cite{jin2014hand}. However, the field witnessed a surge in popularity and progress with the rise of deep learning around 2015, particularly convolutional neural networks (CNNs) adapted for processing point clouds and voxelized data \cite{o2015introduction}. Architectures like VoxelNet (2017) \cite{zhou2018voxelnet} and PointNet/PointNet++ (2017)  \cite{qi2017pointnet} marked turning points by directly learning from raw 3D data without requiring complex preprocessing. These innovations enabled more accurate and scalable 3D detection systems, fueling research in autonomous driving, augmented reality, and robotic manipulation. The growing demand for spatial awareness and environmental understanding in real-world scenarios continues to push the boundaries of 3D object detection, making it a critical component in the advancement of intelligent systems and multimodal AI. Its popularity in academic and industrial research remains strong, with continued development of benchmarks, datasets, and novel architectures. All major models and their processing strategies across with neural networks are presented in Table \ref{tab:3D_object_detection_models_3col}. 

\begin{table}[htbp]
\centering
\caption{Timeline and classification of Traditional Methods 3D object detection (with convolution neural networks)}
\scriptsize
\begin{tabular}{|p{3.5 cm}|p{1.7cm}|p{3.5cm}|}
\hline
\textbf{Model (Release Date) \& Reference} & \textbf{Type} & \textbf{Processing Method} \\
\hline
VoxelNet  \cite{zhou2018voxelnet} & Non-Attention & Voxel-wise \\
PointNet  \cite{qi2017pointnet} & Non-Attention & Point-wise \\
MV3D  \cite{chen2017multi} & Non-Attention & ROI-wise (Multi-view Fusion) \\
\hline
SECOND  \cite{yan2018second} & Non-Attention & Voxel-wise \\
Frustum PointNet   \cite{qi2018frustum} & Non-Attention & Point-wise (RGB-D Frustum Proposal) \\
\hline
MV3D  \cite{chen2017multi}& Non-Attention & ROI-wise \\
AVOD \cite{ku2018joint} & Non-Attention & ROI-wise \\
PointFusion \cite{xu2018pointfusion} & Non-Attention & Point-wise \\
MVX-Net \cite{sindagi2019mvx}& Attention-Based & Voxel-wise \\
\hline
PointPainting \cite{vora2020pointpainting} & Attention-Based & Point-wise \\
3D-CVF \cite{yoo20203d} & Attention-Based & Point-wise \\
EPNet \cite{huang2020epnet} & Attention-Based  & Point-wise \\
LiDAR-RCNN \cite{li2021lidar}& 	Non-Attention	&  ROI-wise (LiDAR-Camera Fusion)\\
DepthFusionNet \cite{shivakumar2019dfusenet}& 	Non-Attention  & 	Voxel-wise (Depth-Guided Fusion)\\
PointAugment \cite{li2020pointaugment}&  Attention-Based  & 	Point-wise (Data Augmentation)\\
Stereo3DNet \cite{chen2020dsgn}	&  Attention-Based   & 	Monocular (Stereo Vision)\\
SVGA Net VoxelGraph \cite{he2022svga}	&  Non-Attention	&  Voxel-wise (Graph Neural Networks)\\
\hline
FusionPainting \cite{xu2021fusionpainting} & Attention-Based & Voxel-wise \\
FusionTransformer cite{fent2024dpft}& Attention-Based	& Hybrid (LiDAR + Camera + Radar)\\
CrossModal3D  \cite{jaritz2022cross} &	Attention-Based	& Multi-Modal Distillation\\
SparseVoxelNet  \cite{chen2023voxelnext} &	Attention-Based	& Sparse Voxel-wise\\
DepthSparse  \cite{wu2022sparse} &	Attention-Based &	Monocular (Depth-Aware Sparsity)\\
\hline
DeepFusion  \cite{li2022deepfusion} & Attention-Based & Voxel-wise \\
TransFusion   \cite{bai2022transfusion}& Attention-Based & ROI-wise \\
UVTR  \cite{li2022unifying} & Attention-Based & Voxel-wise \\
Deep Interaction  \cite{yang2022deepinteraction} & Attention-Based & Point-wise \\
LiDARFormer  \cite{zhou2024lidarformer}& Attention-Based &	Voxel-wise + BEV Fusion\\
\hline
CMT  \cite{yan2023cross} & Attention-Based & Voxel-wise \\
BEVFusion  \cite{zhao2024unibevfusion} & Attention-Based & ROI-wise \\
GA-Fusion \cite{li2024gafusion}(Aug 2023) & Attention-Based & Point-wise \\
IS-Fusion  \cite{yin2024fusion} & Attention-Based & Voxel-wise \\
Uni3D  \cite{zhang2023uni3d}& Attention-Based & Multi-dataset \\
VCD  \cite{huang2023leveraging} & Attention-Based & Multi-modal Distillation \\
Vista \cite{deng2022vista} 	& Attention-Based	&  cross-view spatial attention\\
VoxelFusionNet  \cite{Song2023} &	Attention-Based	& Voxel-wise + Multi-modal\\
PointDiffusion   \cite{chen2024diffubox}&	Diffusion-Based	& Point-wise (LiDAR Denoising)\\
BEVDet++   \cite{huang2021bevdet}&	Attention-Based & 	BEV Grids + ROI-wise\\
FusionFormer  \cite{hu2023fusionformer}&	Attention-Based	& Hybrid (Voxel + Point)\\
\hline
3DiffTection  \cite{xu20243difftection} & Diffusion-Based & Single-image \\
OVM3D-Det  \cite{huang2024training}& Attention-Based & Monocular (Pseudo-LiDAR) \\
3D-SparseCNN 	\cite{shi2022srcn3d}& Attention-Based &	Sparse Voxel-wise\\
MonoDepth3D 	\cite{wang2022monocular}& Attention-Based	& Monocular (Self-Supervised)\\
Panoptic3DNet  \cite{lee2024panopticus} &	Attention-Based	& Voxel-wise + Panoptic Segmentation\\
HyperVoxel  \cite{noh2021hvpr}&	Attention-Based	& Hypernetwork-Driven Voxel-wise\\
\hline
SparseVoxFormer  \cite{son2025sparsevoxformer}& Attention-Based & Voxel-wise \\
Omni3D+ \cite{brazil2023omni3d}& Attention-Based & Benchmark-driven \\
PillarFocusNet  \cite{gao2025pillarfocusnet} & Attention-Based & Pillar-based \\
UniDet3D  \cite{kolodiazhnyi2025unidet3d} & Attention-Based & Voxel-wise \\
MonoTAKD  \cite{liu2024monotakd} & Attention-Based & Monocular (KD) \\
\hline
\end{tabular}
\label{tab:3D_object_detection_models_3col}
\end{table}

Although neural network-based processing methods such as VoxelNet and PointNet have shown promising results, they still face several inherent limitations. These limitations are outlined in the following points:
\begin{itemize}
    \item \textbf{Limitations in Early Fusion and ROI-Based Models}:  Early 3D object detection models such as MV3D (2017) and AVOD (2019) adopted ROI-wise processing strategies based on hand-engineered features and multi-view fusion \cite{chen2017multi, yuan2023attention}. A primary challenge for these models was inefficient spatial reasoning, as the 2D region proposals often failed to align accurately with 3D geometry, especially in occluded or cluttered scenes \cite{chen2017spatial}. Their performance heavily depended on camera calibration and multi-view consistency, which degraded in dynamic or sensor-imprecise environments. PointFusion (2019) aimed to mitigate some of these issues using point-wise fusion of RGB and depth features, yet it struggled with misalignment in feature spaces and lacked a unified mechanism for robust cross-modal attention \cite{xu2018pointfusion}. Likewise, LiDAR-RCNN (2020) introduced camera-LiDAR fusion but still relied on handcrafted heuristics and lacked contextual integration, resulting in limited generalization to novel scenes or weather conditions \cite{li2021lidar}. Furthermore, all these models required rigid sensor setups and suffered from heavy computation when scaling to larger scenes or datasets. In general, these models were constrained by low adaptability, poor cross-modal coordination, and dependence on external proposal generation modules, which limited their scalability and real-time deployment in dynamic environments like autonomous driving.
    
    \item \textbf{Voxelization Bottlenecks and Sparsity :} Voxel-based approaches such as VoxelNet (2017) \cite{sindagi2019mvx} and SECOND (2018) \cite{yan2018second}represented a significant shift by enabling 3D CNNs to directly learn from LiDAR data. While these models improved detection accuracy by capturing geometric structure in 3D, voxelization introduced several key limitations. First, quantization errors and loss of fine-grained information were common due to fixed grid resolutions, which impaired performance on small or thin objects. Second, voxel grids were inherently sparse, yet early models processed them densely, leading to computational inefficiency. VoxelGraph (2020) attempted to exploit voxel sparsity using graph neural networks \cite{lu2023voxel}, but this added architectural complexity and introduced optimization instability in training. MonoDFNet (2025) used depth-guided fusion to improve voxel representation, but it was highly dependent on accurate depth maps, which are often noisy in low-light or reflective scenes \cite{gao2025monodfnet}. Across these voxel-centric models, a common bottleneck was the trade-off between spatial resolution and efficiency. Moreover, since they lacked attention mechanisms, these models struggled with context modeling across long spatial ranges, making them less effective in scenes with multiple object interactions or partial occlusions.

    \item \textbf{Challenges in Point-Wise and Frustum-Based Models:} Point-based networks, especially PointNet (2017) and its successor PointNet++, pioneered direct learning on raw 3D point clouds \cite{qi2017pointnet}. These models eliminated the need for voxelization \cite{zhou2020end}, preserving spatial fidelity \cite{zhu2022vpfnet}, but encountered critical limitations. PointNet's architecture was order-invariant but spatially myopic, meaning it could not model local context effectively. PointNet++ improved this by introducing hierarchical grouping, but still suffered from non-uniform sampling issues and high sensitivity to point cloud density \cite{sheshappanavar2020novel, li20213d}. Frustum PointNet (2018) \cite{qi2018frustum}combined 2D object proposals with 3D point processing, achieving decent results in constrained environments, but was highly dependent on the accuracy of 2D detectors, making it unsuitable for autonomous applications with noisy camera inputs. EPNet (2020) \cite{huang2020epnet} and PointAugment (2020) \cite{li2020pointaugment} incorporated attention and augmentation strategies to enhance point-wise reasoning, but their reliance on manually tuned fusion mechanisms and limited receptive fields restricted their generalization in cluttered or unstructured scenes. These models also lacked dynamic adaptability to unseen object classes, as their training was anchored in closed-set object categories. Collectively, point-based models before 2022 were constrained by scalability, context integration, and real-time performance bottlenecks \cite{sapkota2025multimodal}, particularly in large-scale outdoor environments with varying lighting and weather conditions.

    \item \textbf{Sensor Dependence, Monocular Instability, and Fusion Complexity:} Multi-modal fusion models like FusionTransformer (2021) and CrossModal3D (2021) \cite{zhao2022crossmodal} aimed to improve robustness by integrating data from LiDAR, cameras, and even radar. However, this came with significant fusion complexity and synchronization challenges. Aligning features across modalities with different spatial and temporal resolutions required careful calibration and preprocessing. Likewise, DepthSparse (2021) \cite{fan2022fully} relied on stereo vision and monocular depth estimation, which are inherently unstable under varying lighting or textureless surfaces, leading to depth estimation errors and false positives. FusionPainting (2021) and SparseVoxelNet (2021) attempted to incorporate sparse attention and temporal fusion but often required large computational resources and showed degraded performance on long-range or partially occluded objects. Another limitation was the lack of semantic reasoning capability—fusion mechanisms were primarily geometric or photometric, unable to leverage textual or contextual cues from the environment. These models also exhibited rigid data modality requirements, limiting their deployment in scenarios where not all sensors are available or calibrated. The limitations in adaptive fusion, computational load, and semantic understanding in these pre-2022 models created a clear need for more flexible, unified, and context-aware approaches—gaps later addressed by Vision-Language models.
\end{itemize}

\subsubsection{3D Object Detection with Vision Language Models:}
Building on early innovations in 3D object detection, the period from 2019 to 2025 marked a transformative shift in methodology, driven by advances in multimodal learning and the integration of VLMs. Initially, models such as MV3D\cite{chen2017multi}, AVOD, and PointFusion relied on non-attention mechanisms and simple ROI or point-wise processing. However, with the rise of attention-based frameworks like MVX-Net and PointPainting in 2019–2020, 3D detection began leveraging cross-modal reasoning and spatial alignment through transformers and multi-modal fusion. The evolution accelerated post-2022, with voxel-wise architectures such as UVTR and DeepFusion optimizing spatial granularity and computation. Notably, recent years have seen the emergence of VLM-powered 3D models, which integrate textual context and visual perception. Models like PaLM-E \cite{driess2023palm}, BLIP-2 \cite{li2023blip}, InternVL, and CogVLM pioneered the use of vision-language embedding alignment, enabling instruction-following, scene understanding, and open-vocabulary detection. By 2024–2025, this trend matured with models like VLM3D-Guide, Instruct3D, and OmniVLM3D, incorporating natural language prompts for more adaptive and interactive 3D understanding. These approaches reflect a broader movement toward sensor-agnostic, context-aware, and language-informed 3D perception systems.  All state-of-the-art VLMs-based 3D object detection are presented in Table \ref{tab:3D_object_detection_models_VLMs}.

\begin{table}[htbp]
\centering
\caption{Timeline and classification of 3D object detection with Vision-Language Models}
\scriptsize
\begin{tabular}{|p{3 cm}|p{1.7cm}|p{3.3cm}|}
\hline
\textbf{Model (Release Date) \& Reference} & \textbf{Type} & \textbf{Processing Method} \\
\hline

PaLM-E  \cite{driess2023palm} &		Attention-Based &		Embodied 3D Perception\\
LLaVA-1.5  \cite{zhu2024llava} &	Attention-Based	& Multimodal Visual Grounding\\
BLIP-2 \cite{li2023blip}	 &	 Encoder-Decoder	 &	 Image-Text 3D Localization\\
InternVL  \cite{zhu2025internvl3}	 &	 Encoder-Decoder	 &	 Visual-Language Spatial Reasoning\\
CogVLM 	\cite{wang2024cogvlm, tian2024hpe} &	 Attention-Based  &	 	Vision-Language Spatial Alignment\\
CLIP3D-Det 	\cite{hegde2023clip}& Attention-Based &	Monocular (CLIP Feature Fusion)\\
Instruct3D  \cite{kamata2023instruct}	& Attention-Based	& ROI-wise (Instruction Tuning)\\
M3D-LaMed  \cite{bai2024m3d} & Encoder-Decoder & 3D Medical Image Understanding via Multi-modal Instruction Tuning \\
Qwen2-VL  \cite{wang2024qwen2}	&  Decoder-Only & 	Multimodal 3D Scene Understanding \\
Find n’ Propagate  \cite{etchegaray2024find} & Hybrid Top-down/Bottom-up & Open-Vocabulary 3D Detection using Frustum Search, Cross-Modal Propagation, and Remote Simulation \\
\vspace{0.05cm}
OWL-ViT  \href{https://huggingface.co/docs/transformers/en/model_doc/owlvit}{ Link to paper }	& Transformer + CLIP	&Open-Vocabulary, Vision-Language Embedding Alignment\\
\vspace{0.05cm}
OmniVLM3D \cite{chen2024omnivlm}	& Attention-Based &	Multi-Dataset + Language Alignment\\
\vspace{0.05cm}
Talk2PC (2025) \cite{guan2025talk2pc} & Prompt-Guided Sensor Fusion & LiDAR-Radar 3D Visual Grounding with Multi-Stage Cross-Modal Attention and Graph Fusion \\
\vspace{0.05cm}
OpenScene (2023) \cite{peng2023openscene} & CLIP-Based Zero-Shot & Open-Vocabulary 3D Scene Understanding via CLIP Feature Embedding \\
\vspace{0.05cm}
Language-Grounded Indoor 3D Semantic Segmentation (2022) \cite{rozenberszki2022language} & Language-Guided Segmentation & Text-Embedded Feature Alignment for Large-Vocabulary 3D Segmentation \\
\vspace{0.05cm}
3D-LLM (2023) \cite{hong20233d} & Multi-Task 3D Reasoning & 3D-Aware Prompting with Multi-View Rendering and VLM Integration for Broad 3D Understanding \\
\vspace{0.05cm}
Text2Loc (2024) \cite{xia2024text2loc} & Language-Guided Localization & Coarse-to-Fine Transformer-Based 3D Localization with Contrastive Learning and Matching-Free Refinement \\
PromptDet \cite{guo2025promptdet} & Attention-Based & Voxel-wise \\
\vspace{0.05cm}
OpenMask3D (2023) \cite{takmaz2023openmask3d} & Open-Vocabulary Segmentation & Zero-Shot 3D Instance Masking via Multi-View CLIP Feature Fusion \\
\vspace{0.05cm}
3D-Grounded VLM Framework (2025) \cite{tang20253d} & Task Planning & Automated 2D-to-3D Prompt Synthesis with SLM-Supervised VLM Reasoning \\
\vspace{0.05cm}
OV-SCAN: Semantically Consistent Alignment for Novel Object Discovery (2025) \cite{chow2025ov} & Cross-Modal Alignment & Open-Vocabulary 3D Object Detection with Semantic Consistency \\
\vspace{0.05cm}
Language-Driven Active Learning for Diverse Open-Set 3D Object Detection (2025) \cite{greer2025language} & Active Learning & Vision-Language Embedding Diversity for Novel Object Detection \\
\vspace{0.05cm}
ULIP: Learning a Unified Representation of Language, Images, and Point Clouds for 3D Understanding (2023) \cite{xue2023ulip} & Multimodal Pre-Training & Unified Representation of 3D Point Clouds, Images, and Language for 3D Classification \\
\vspace{0.05cm}
OpenShape: Scaling Up 3D Shape Representation Towards Open-World Understanding (2023) \cite{liu2023openshape} & Multimodal Contrastive Learning & Open-World 3D Shape Understanding via Multi-Modal Representation Alignment \\

\hline
\end{tabular}
\label{tab:3D_object_detection_models_VLMs}
\end{table}

VLMs have recently emerged as transformative solutions to longstanding limitations in traditional 3D object detection pipelines. Unlike earlier models that struggled with rigid sensor dependencies, modality-specific architectures, and limited semantic reasoning, VLM-based systems integrate textual context and visual features within a unified, cross-modal framework \cite{luo2024delving, zhang2024vision}. This enables a richer understanding of scenes, where detection is guided not only by geometry but also by high-level language priors. Models like PaLM-E (2023) and CogVLM (2023) \cite{wang2024cogvlm} exemplify this shift, supporting embodied perception by aligning linguistic commands or descriptions with visual and spatial input for downstream 3D tasks. Similarly, BLIP-2 (2023) and InternVL (2023) introduced powerful encoder-decoder architectures that enhance 3D object grounding through multimodal fusion, improving performance in open-vocabulary and zero-shot settings. In 2024, models like TextVoxelNet and Instruct3D further extended these capabilities by incorporating language-guided voxel reasoning and instruction-tuned 3D detection, enabling more controllable and human-aligned outputs \cite{kamata2023instruct}. These VLMs address key challenges such as poor generalization, lack of semantic understanding, and the need for multi-sensor alignment by offering flexible, text-conditioned object proposals and scene interpretations. Furthermore, recent VLMs like ZeroVL3D (2025) push boundaries in zero-shot 3D detection, functioning effectively in novel environments without retraining. Collectively, these models mark a paradigm shift toward more adaptive, interpretable, and semantically enriched 3D perception systems. The integration of vision-language pretraining, cross-modal attention, and textual prompting positions VLMs as a promising foundation for the next generation of 3D object detection. The next sections will delve deeper into their mechanisms, capabilities, and limitations.

\begin{figure*}[h!]
  \centering
  \includegraphics[width=0.7\linewidth]{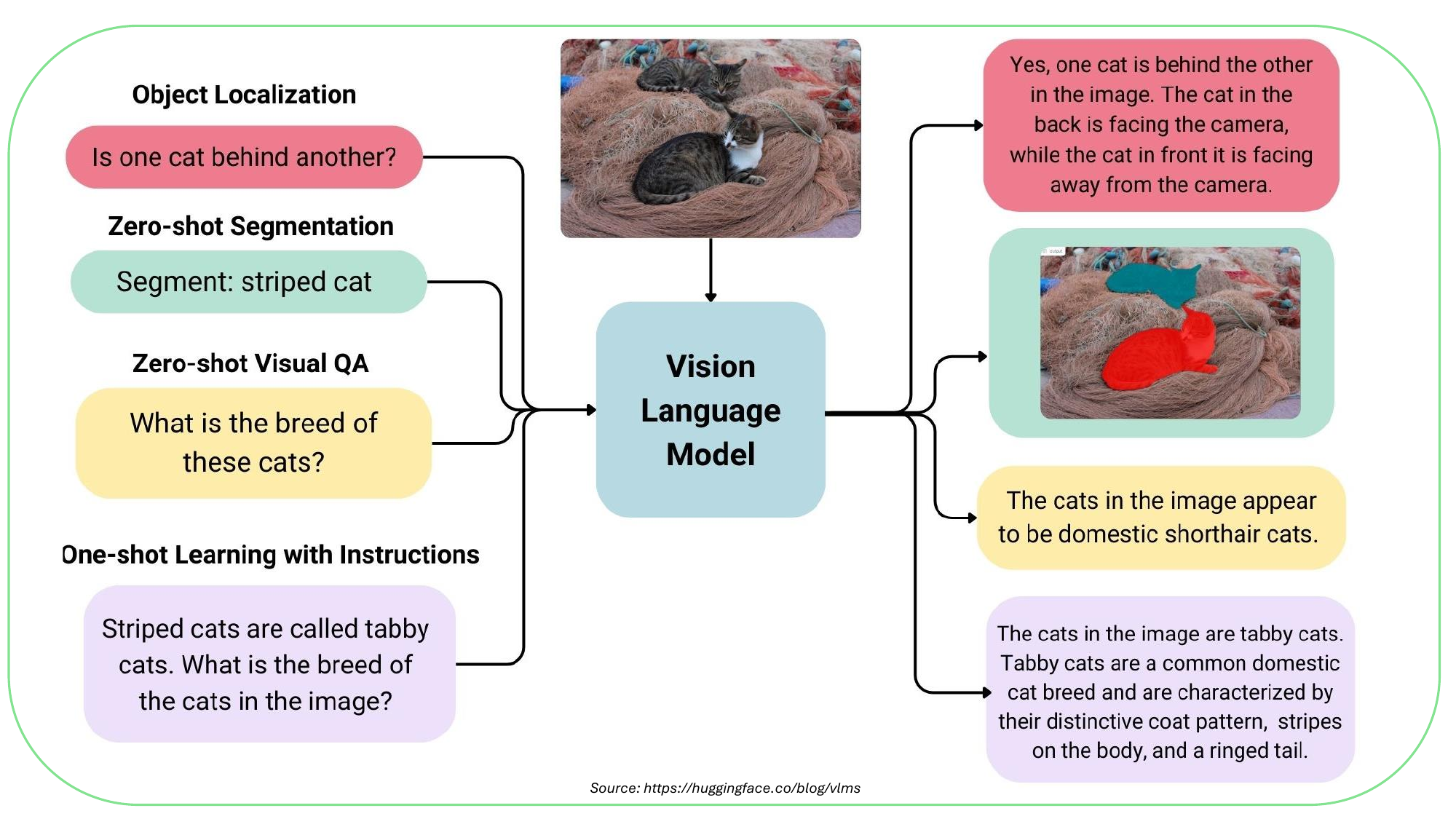}
  \caption{Illustration of a Vision-Language Model (VLM) performing multimodal reasoning—detecting, segmenting, and describing objects from images using textual prompts, demonstrating open-vocabulary 3D detection and contextual understanding.}
  \Description{Image2}
  \label{fig:vlmhuggingface}
\end{figure*}

The VLMs serve as the backbone for a variety of multimodal tasks, including object detection, image captioning, visual question answering (VQA), and image segmentation \cite{li2025benchmark}. Their ability to generalize across diverse visual inputs and interpret complex textual prompts makes them highly adaptable across domains \cite{du2022learning}. At the core of a typical VLM are three primary components: an image encoder, a multimodal fusion module, and a text decoder. The image encoder—commonly a Vision Transformer orCNN—encodes visual inputs into rich embeddings. These are then aligned with textual embeddings through cross-attention mechanisms or projection layers within the fusion module, facilitating meaningful cross-modal interaction \cite{khan2022transformers, li2023transformer}. The final component, a LLM, decodes this integrated representation to produce coherent and contextually appropriate language outputs. VLMs go beyond traditional visual systems by enabling open-vocabulary object detection, where objects can be identified based on arbitrary language queries rather than predefined class labels. This is accomplished by mapping both image regions and text into a shared embedding space, which allows flexible and context-aware reasoning. As illustrated in Figure \ref{fig:vlmhuggingface}, a VLM analyzing an image of two cats can determine spatial relationships (“Is one cat behind the other?”), segment visual entities based on descriptive prompts (“Segment: striped cat”), and answer contextual questions (“What is the breed of these cats?”). Furthermore, it can adapt through instruction, such as learning that “striped cats are called tabby cats,” and applying that information to reclassify visual inputs. This example underscores the comprehensive reasoning capabilities of VLMs, showcasing their role as dynamic interpreters of multimodal content.

\subsection{ Architecture and Foundations of VLMs for 3D Object Detection}
\subsubsection{Pretraining and Fine-tuning Structure of VLMs for Object Detection:}
\begin{figure*}[h!]
  \centering
  \includegraphics[width=0.7\linewidth]{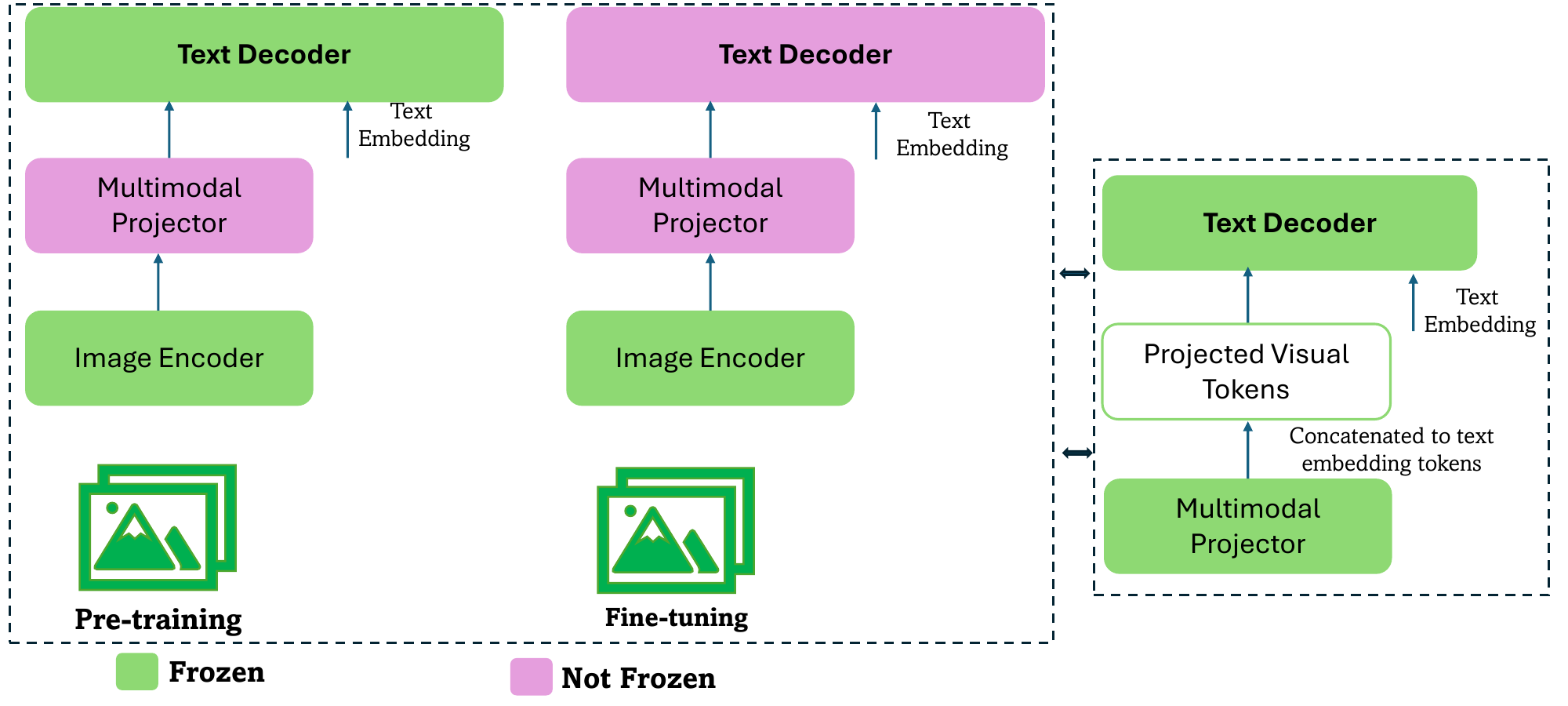}
  \caption{Overview of VLM architecture showing pretraining and fine-tuning stages. Visual embeddings from an image encoder are projected into a shared space and concatenated with text tokens. During pretraining, only the projector is trained to align visual and textual representations. In fine-tuning, the decoder is adapted to downstream tasks like object detection. This modular design supports flexible multimodal learning, enabling VLMs to generalize across vision-language tasks such as grounding, captioning, and 3D object detection.}
  \Description{Imageadding}
  \label{fig:VLMsstructure}
\end{figure*}

The architectural pipeline of VLMs, as illustrated in Figure \ref{fig:VLMsstructure}, follows a sequential design with distinct pretraining and fine-tuning phases. During pretraining, the model learns cross-modal alignment through a three-stage framework: an image encoder (e.g., CLIP-ViT) processes input images into visual embeddings, a multimodal projector (e.g., dense neural networks) maps these embeddings into the text token space, and a text decoder (e.g., Vicuna) generates captions or answers autoregressively \cite{huang2023visual, li2024multimodal}. The core technical challenge lies in unifying image and text representations. To achieve this, visual tokens are projected into the decoder’s embedding space and concatenated with text tokens, forming a fused input sequence \cite{chen2022unified}. For example, LLaVA employs a CLIP-based encoder to extract image features, which are linearly projected into Vicuna’s token dimension \cite{zhu2024llava}. These projected visual tokens are prepended to text embeddings (e.g., questions or prompts), enabling the decoder to process multimodal inputs seamlessly.

\begin{figure*}[h!]
  \centering
  \includegraphics[width=0.98\linewidth]{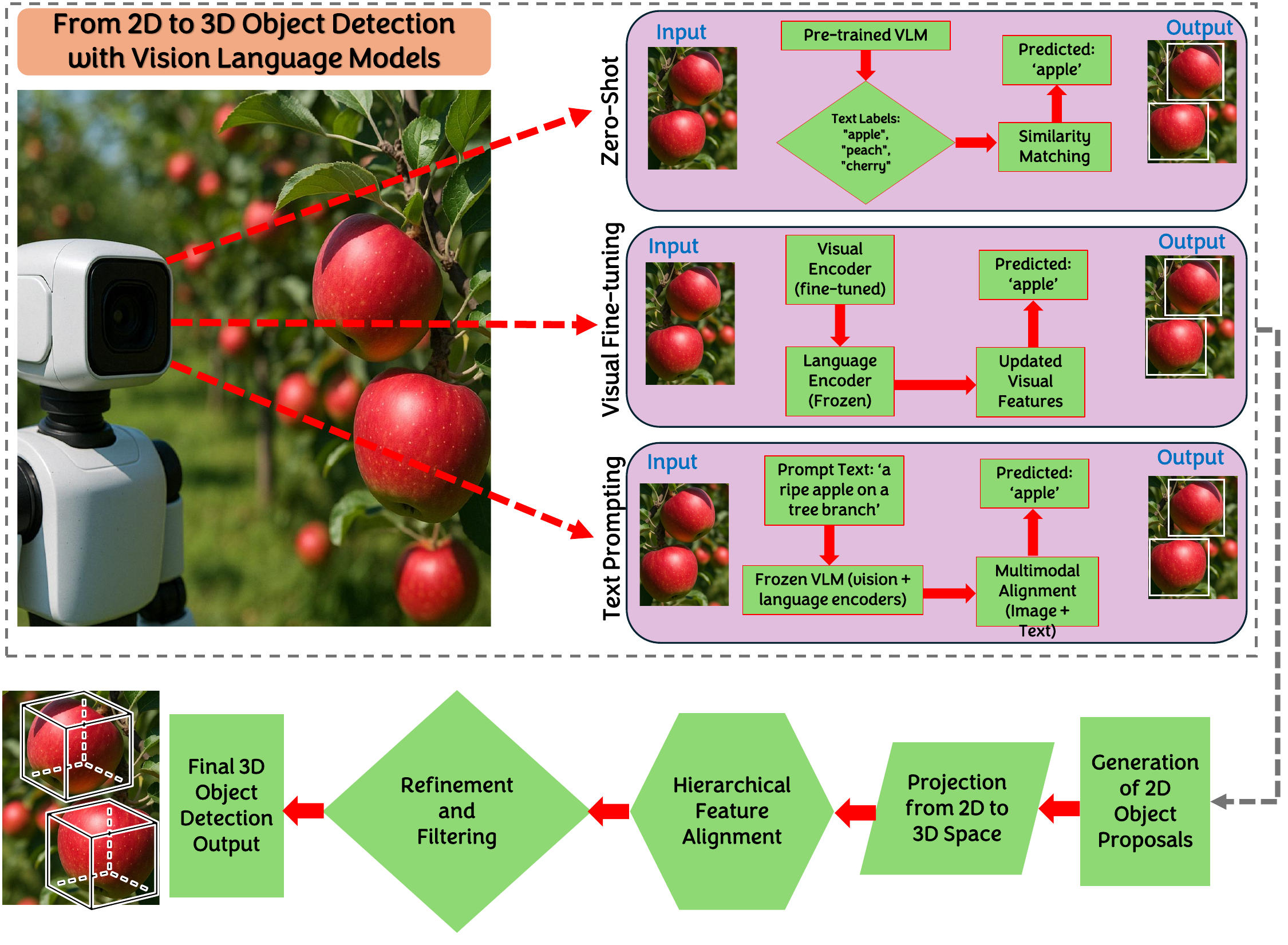}
  \caption{Illustration of 2D and 3D object detection using VLMs in a commercial orchard setting. The robot uses VLMs to perform 2D detection by aligning visual features with textual prompts, enabling zero-shot object identification. These 2D proposals are projected into 3D space using depth data to generate volumetric bounding boxes. Cross-modal fusion with LiDAR or depth inputs refines object localization, allowing context-aware reasoning about apple position and ripeness. This pipeline highlights how VLMs bridge semantic understanding and geometric reasoning for real-world agricultural tasks, enabling robots to detect and interact with fruit in complex, dynamic environments.}
  \Description{Image }
  \label{fig:2dto2dVLMs}
\end{figure*}

Pretraining strategies vary, but a common approach involves two stages. First, the multimodal projector is trained to align visual and textual features while keeping the image encoder and text decoder frozen. For instance, LLaVA leverages GPT-4 to synthesize instruction-response pairs from image-caption datasets: images and captions are fed to GPT-4 to generate contextual questions, which serve as training data \cite{zhu2024llava}. The projector is optimized using cross-entropy loss to align visual embeddings with the decoder’s expected text space. In the second stage, the text decoder is unfrozen, and both the projector and decoder are jointly fine-tuned on task-specific data (e.g., visual QA pairs), refining cross-modal reasoning.

During fine-tuning, the pretrained image encoder and projector remain fixed or lightly updated, while the text decoder is adapted to downstream tasks. For detection or grounding, textual instructions (e.g., “Output bounding boxes for all cars”) are concatenated with projected visual tokens, conditioning the decoder to generate structured outputs like coordinates or masks. This phased training ensures computational efficiency—only a fraction of parameters (projector and decoder) are updated—while preserving the encoder’s generalizable visual features.

Key innovations include the use of lightweight projectors (e.g., linear layers) for parameter efficiency and spatial-aware architectures that preserve pixel-level details for localization tasks. The modular design—freezing encoders, training projectors, and adapting decoders—enables VLMs to scale across diverse applications, from captioning to 3D detection, while maintaining robust zero-shot capabilities.

\textbf{2D Object Detection with VLMs:}
VLMs detect objects in 2D environments through a multimodal process that combines visual feature extraction and semantic alignment (Fig. \ref{fig:2dto2dVLMs}). First, an image encoder (e.g., Vision Transformer) processes the input image into a grid of spatial embeddings, capturing hierarchical features such as edges, textures, and object parts. These embeddings are projected into a joint vision-language space via a multimodal projector (e.g., linear layers), aligning them with textual token embeddings\cite{zang2025contextual, jain2024vcoder}. The text decoder then interprets these fused representations to generate bounding boxes or segmentation masks. Three primary strategies govern this process:

\begin{itemize}
    \item \textbf{Zero-Shot Prediction}: The model matches visual features to textual labels (e.g., "car") without fine-tuning, leveraging pretrained semantic relationships.
    \item \textbf{Visual Fine-Tuning}: The visual encoder is adapted to domain-specific data (e.g., urban scenes) to improve detection accuracy for rare objects.
    \item \textbf{Text Prompting}: Descriptive prompts (e.g., "a red truck near the curb") guide the decoder to focus on contextual details, enhancing localization precision.
\end{itemize}

The output includes 2D bounding boxes and class labels, generated autoregressively by correlating textual queries with activated visual regions. This process excels in generalizability but lacks depth-aware reasoning.

\textbf{3D Object Detection with VLMs:}
Extending VLMs to 3D detection involves a multi-stage pipeline (Fig. \ref{fig:2dto2dVLMs}), where 2D proposals are fused with 3D geometric data to predict spatially grounded objects. The steps are as follows:

\begin{enumerate}
    \item \textbf{2D Object Proposals}: The VLM first generates 2D bounding boxes and class labels using its vision-language alignment capabilities. For example, it identifies "pedestrian" regions in an image through zero-shot prediction or text prompts like "highlight all pedestrians."
    
    \item \textbf{2D-to-3D Projection}: The 2D proposals are projected into 3D space using depth maps or LiDAR-Camera calibration matrices. Each 2D region is mapped to a frustum in 3D, creating candidate 3D volumes. For instance, a 2D "car" bounding box is extruded into a 3D cuboid based on estimated depth.
    
    \item \textbf{Hierarchical Feature Alignment}: The VLM aligns 2D visual tokens (from the image encoder) with 3D point cloud features (from LiDAR or depth sensors). This is achieved through cross-modal attention mechanisms, where 3D points within the projected frustum attend to relevant 2D regions. For example, points in a "tree" frustum aggregate features from image patches showing leaves or branches.
    
    \item \textbf{Refinement and Filtering}: The initial 3D proposals are refined using geometric constraints (e.g., size priors) and semantic feedback from the VLM. The text decoder validates proposals by correlating 3D features with textual queries (e.g., "a sedan with dimensions 4.5m x 1.8m"). Outliers are filtered using confidence scores derived from vision-language similarity.
\end{enumerate}


\subsubsection{Visualization of 3D Detection with VLMs:}
Recent advancements in 3D object detection have been significantly influenced by the integration of VLMs, enabling models to understand and reason about 3D scenes in a more versatile and semantically grounded manner. In a study by Chen et al. \cite{chen2024spatialvlm}, the SpatialVLM model demonstrates how VLMs can leverage large-scale vision-language pretraining and spatial question-answering tasks to implicitly learn 3D relationships from monocular images, bypassing the need for traditional annotated 3D data or depth sensors. This innovative approach allows for both quantitative and qualitative reasoning in 3D environments. However, its reliance on synthetic question-answering data may limit real-world applicability. Further developments by Hong et al. \cite{hong20233d} and Jiao et al. \cite{jiao2024unlocking} push the boundaries of VLM integration by combining multi-view rendered point cloud data with large language models. The 3D-LLM framework \cite{hong20233d} and hierarchical alignment methods in \cite{jiao2024unlocking} enable zero-shot and few-shot generalization for novel object detection tasks, addressing limitations of traditional methods that often require extensive annotated datasets. Additionally, models like CoDA \cite{cao2023coda} and MSSG \cite{cheng2023language} showcase how combining geometry, semantics, and language can enhance open-vocabulary detection and improve interpretability, marking a transformative shift in 3D perception research.

In a recent study by Brazil et al. \cite{brazil2023omni3d} introduces OMNI3D, a large-scale benchmark for 3D object detection that re-purposes existing datasets to address the limitations of smaller, domain-specific datasets in 3D recognition. OMNI3D consists of 234,000 images and over 3 million annotated instances across 98 object categories, making it 20 times larger than existing 3D benchmarks like KITTI \cite{geiger2013vision} and SUN RGB-D \cite{song2015sun}. A key challenge in 3D object detection is the diversity in camera intrinsics and object types. To overcome this, the authors propose Cube R-CNN  \cite{piekenbrinck2024rgb}, a unified model designed to generalize across varying camera and scene types. Cube R-CNN predicts 3D object locations, sizes, and rotations from a single image, leveraging a novel approach called virtual depth to mitigate scale-depth ambiguity caused by variations in camera focal lengths. This model outperforms previous methods, such as ImVoxelNet and GUPNet, on both urban and indoor benchmarks, demonstrating strong generalization across different domains. OMNI3D's large scale and diversity also enable improvements in single-dataset performance and accelerate learning on smaller datasets through pre-training.

\begin{figure*}[h!]
  \centering
  \includegraphics[width=0.9\linewidth]{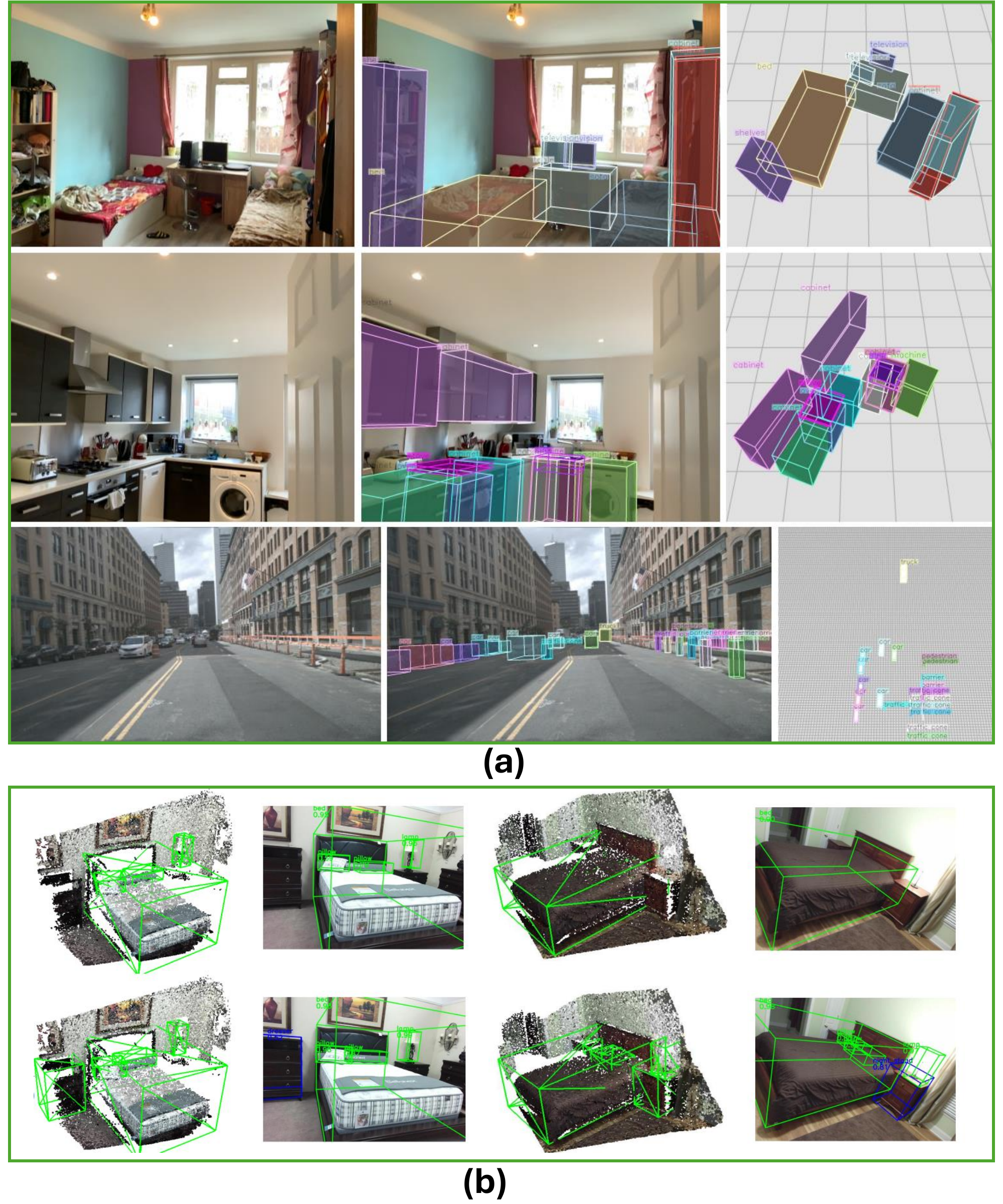}
  \caption{ Examples of 3D Object detection with VLMs:(a)  Cube R-CNN with VLM-based 3D object detection on OMNI3D, showcasing predictions in a room, kitchen, and traffic scenes with 3D bounding boxes.\cite{brazil2023omni3d} ; b)  Illustrates CoDA's 3D detection process \cite{cao2023coda}, showcasing its ability to localize and classify novel objects with high accuracy in diverse scenes.}
  \Description{Image }
  \label{fig:detection1}
\end{figure*}

Figure \ref{fig:detection1}a demonstrates the performance of Cube R-CNN on the OMNI3D test set, showcasing its ability to predict 3D objects across diverse environments. The upper section of the figure displays the detection of objects in a room, such as a bed, table, and cupboard, accurately predicted in 3D space. These predictions are overlaid on the input image, providing a clear visualization of how Cube R-CNN recognizes and locates these objects within the scene \cite{piekenbrinck2024rgb}. The 3D bounding boxes are also shown from a top-down perspective, where the base is composed of 1m×1m tiles, offering an additional view of the spatial arrangement of these objects. The middle part of the figure focuses on the detection of kitchen items, including washing machines and racks, with 3D bounding boxes accurately identifying their positions and sizes in the image. This demonstrates the model's versatility in handling diverse object categories. In the lower section of Figure \ref{fig:detection1}a, Cube R-CNN performs 3D detection on a traffic road scene, identifying vehicles and pedestrians in a dynamic, outdoor environment. This variety of scenes, from indoor to outdoor, highlights Cube R-CNN's robust ability to generalize across different domains, making it a powerful tool for large-scale 3D object detection.

Likewise, CoDA \cite{cao2023coda} introduces a novel approach to open-vocabulary 3D object detection (OV-3DDet), addressing the challenges of localizing and classifying novel objects in 3D scenes with limited base categories. The paper proposes a unified framework that integrates two key components: the 3D Novel Object Discovery (3D-NOD) strategy and a cross-modal alignment module. The 3D-NOD strategy leverages both 3D geometry priors and 2D semantic priors from the CLIP model to localize novel objects, creating pseudo-labels for objects not seen during training. To improve classification, CoDA employs the Discovery-Driven Cross-Modal Alignment (DCMA), which aligns 3D object features with 2D image/text features. This alignment consists of class-agnostic distillation and class-specific contrastive learning, which iteratively enhance the model’s ability to detect novel objects by leveraging the discovered novel object boxes. The framework works without relying on pre-annotated 2D open-vocabulary detection models, using a CLIP-based vision-language model to transfer rich open-world knowledge into 3D object detection. CoDA’s iterative process of novel object discovery and feature alignment significantly improves performance, outperforming the best alternative methods by 80\% in mean Average Precision (mAP) on benchmark datasets like SUN-RGBD and ScanNet. This method demonstrates the ability to localize and classify novel objects by continuously expanding a pool of discovered novel boxes and refining feature alignment, leading to better detection and classification in an open-vocabulary setting.

As shown in Figure \ref{fig:detection1}b, the model demonstrates its strong capability to discover more novel objects from point clouds. For instance, our method successfully identifies the dresser in the first scene and the nightstand in the second scene. This highlights the framework’s ability to effectively localize and classify novel objects that were not part of the base categories. The performance comparison further indicates that our method generally exhibits superior open-vocabulary detection ability on novel objects, outperforming alternative methods in terms of precision and detection accuracy. The figure illustrates the 3D detection process, showcasing how the model identifies these objects with high accuracy in diverse 3D environments. Wang et al. \cite{wang2024g} propose G3-LQ, a model for 3D visual grounding in Embodied AI that integrates geometric and semantic cues using modules like PAGE and Flan-QS. By introducing Poincaré Semantic Alignment loss and evaluating on ScanRefer and Nr3d/Sr3d, G3-LQ achieves superior alignment between text and 3D features, outperforming prior methods.

\begin{figure*}[h!]
  \centering
  \includegraphics[width=0.9\linewidth]{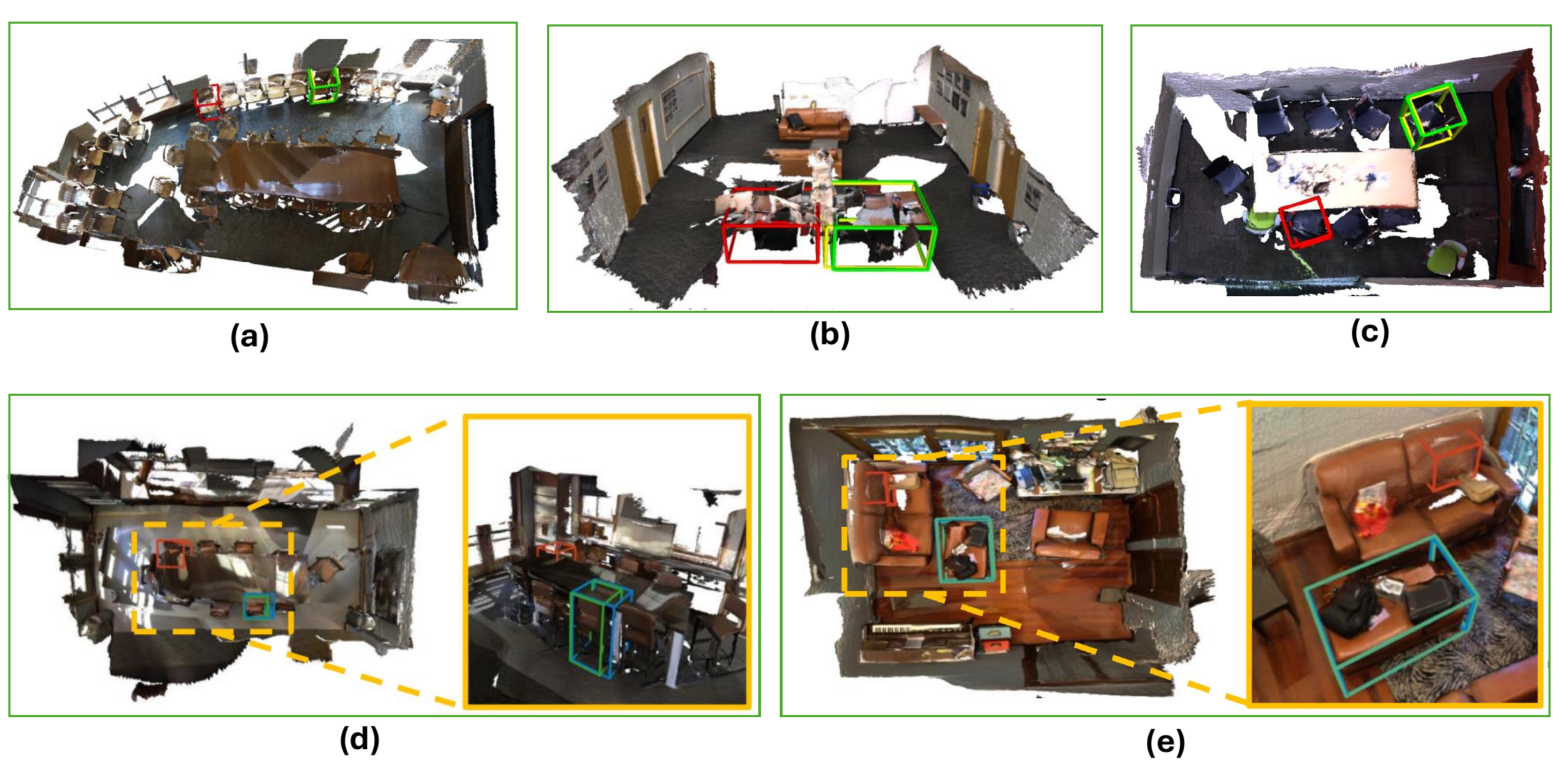}
  \caption{ Examples of 3D Object Detection with VLMs: a) shows 3D detection with VLMs for "this is a chair by the wall, the fifth chair from the right wall" by \cite{wang2024g}; b) illustrates "the left-most table in the center" bby \cite{wang2024g}; c) depicts "a black office chair." by\cite{wang2024g} ; de) \cite{zhang2024vision}}
  \Description{Image }
  \label{fig:detection2}
\end{figure*}

As shown in Figure \ref{fig:detection2}, we present examples of 3D detection proposed by the author in G3-LQ on the ScanRefer dataset. The figure illustrates the model's remarkable grounding performance in interpreting both geometric attributes and complex language queries. In Figure \ref{fig:detection2}a, the query "this is a chair by the wall. it is the fifth chair from the right wall surface" is correctly grounded, showcasing the model’s ability to recognize object positions based on spatial relationships. Figure \ref{fig:detection2}b displays the description "the table is the left-most one in the center of the room. the table is dark orange and has four legs," highlighting G3-LQ’s proficiency in discerning object size, color, and orientation. Additionally, Figure \ref{fig:detection2}c illustrates "it is a black office chair. the black office chair is the fourth office chair on the left side of the table," demonstrating the model’s capacity to effectively parse complex utterances and spatial ordering.

Likwise, Zhang et al. (2024) \cite{zhang2024vision} present 3DVLP, a vision-language pre-training framework tailored for semantic 3D scene understanding. Their goal is to learn a universal embedding that generalizes across downstream tasks like 3D object detection, grounding, and captioning. Unlike prior task-specific models, 3DVLP adopts visual grounding as a proxy task and introduces three object-level modules to improve proposal quality and semantic alignment. First, the Object-level IoU-guided Detection (OID) loss uses DIoU and label smoothing to generate accurate 3D bounding boxes. Second, Object-level Cross-Contrastive alignment (OCC) aligns proposal features with text embeddings by maximizing similarity between matched pairs while pushing mismatches apart. Likewise, Object-level Self-Contrastive learning (OSC) enhances intra-modal discrimination by ensuring distinct proposal features for different objects \cite{zhang2024vision}. All three modules rely on an IoU-based filtering mechanism to define positive and negative training pairs. These components enable the model to handle issues of modality misalignment, overlapping semantic information, and fine-grained object differentiation. During fine-tuning, 3DVLP’s backbone is transferred to task-specific heads, demonstrating strong generalization and competitive performance. Notably, the framework supports efficient fine-tuning, making it highly adaptable.

Figure \ref{fig:detection2}d shows an example 3D detection output from 3DVLP, identifying the object described as “There is a tall chair pulled up to the table in the room. It is the second from the right.” Likewise, Figure \ref{fig:detection2}e presents a 3D visual grounding result where the model accurately detects the brown ottoman positioned in front of a brown sofa. The ottoman is correctly identified along with detailed attributes, including a black backpack, a black duffel bag, and a box of tissues placed on it. This qualitative result underscores the effectiveness of the 3DVLP model trained on the ScanRefer dataset, where 3DVLP achieves state-of-the-art performance with an accuracy of 51.70\% at IoU@0.25 and 40.51\% at IoU@0.5. These findings highlight the model’s superior grounding ability, particularly in complex scenes with multiple, visually similar objects.

\subsection{Strengths and Challenges of 3D Object Detection with VLMs}
Recent advances in 3D object detection via VLMs have shown impressive potential, particularly in zero-shot and open-vocabulary scenarios \cite{sapkotaaobject, sapkota2024zero}. Unlike traditional 3D detectors that depend heavily on large-scale annotated datasets, modern VLM-based approaches harness pretrained 2D vision-language priors (e.g., CLIP, GPT-3/4) to bridge the modality gap between images, point clouds, and language. Notably, PointCLIP V2 \cite{zhu2023pointclip} demonstrates how realistic projection modules and GPT-generated prompts can achieve strong 3D performance without any 3D training data. Similarly, 3DVLP \cite{zhang2024vision} leverages object-level contrastive learning to boost transferability across tasks such as detection, grounding, and captioning. Language-free 3DVG \cite{zhang2024towards} further reduces reliance on human annotations by synthesizing pseudo-language features from multi-view images, achieving effective grounding without textual supervision. Other works like Uni3DL \cite{li2024uni3dl} and CoDA \cite{cao2023coda} focus on architectural unification and cross-modal alignment to enhance scalability and generalization in real-world 3D scenes. However, several challenges remain. Many methods rely on complex training pipelines (e.g., rendering, projection, pseudo-language modeling), increasing computational cost and limiting real-time applicability. Others face generalization issues when exposed to sparse data, unfamiliar object categories, or diverse environments. Furthermore, performance can be sensitive to prompt design, quality of multi-view imagery, or LiDAR calibration accuracy. Some models also depend heavily on pretrained components and handcrafted alignment mechanisms, which may not scale efficiently across datasets or applications.

These recent approaches demonstrate a diverse range of methodologies for tackling 3D object detection using vision-language models. PointCLIP V2 \cite{zhu2023pointclip} adopts a dual-prompting strategy that bridges 3D point clouds with 2D vision-language priors via depth-based projection and language synthesis, allowing for strong zero-shot performance without 3D supervision. Language-free 3DVG \cite{zhang2024towards} replaces text entirely with pseudo-language embeddings derived from multi-view images, reinforcing spatial relations using cross-modality consistency modules. 3DVLP \cite{zhang2024vision} takes a contrastive learning route, defining object-level alignment tasks like OID, OCC, and OSC to build robust multimodal embeddings. Meanwhile, SpatialVLM \cite{chen2024spatialvlm} frames detection as a spatial question-answering task, using large-scale VQA datasets aligned with monocular depth predictions. 3D-LLM \cite{hong20233d} integrates multi-view rendered features directly into LLMs and supplements them with point cloud-based localization modules. Some works, like CoDA \cite{cao2023coda}, focus on open-vocabulary detection by aligning 3D geometry with textual/image priors to discover unseen classes. Others such as MSSG \cite{cheng2023language} introduce novel tasks like 3D referring expression comprehension, merging language grounding with LiDAR-based detection in real-time settings. Unsupervised methods like those by \cite{fruhwirth2024vision} and \cite{najibi2023unsupervised} exploit CLIP-based classification over LiDAR clusters or auto-labeling with vision-language distillation, removing reliance on labeled 3D data. Furthermore, Uni3DL \cite{li2024uni3dl} proposes a unified transformer framework that operates directly on point clouds, dynamically routing tasks across 3D vision-language objectives through a shared token space. These varied methodologies highlight the flexibility of VLMs in adapting to the unique demands of 3D understanding, from geometric projections to task-specific grounding.  Table below summarizes the key contributions, strengths, and limitations of recent VLM-based 3D object detection methods, offering a comparative perspective on this rapidly evolving research landscape.

\textbf{Comparison of VLM-based 3D object detection methods, highlighting strengths, innovations, and key limitations:}

\onecolumn 
\begin{longtable*}{@{}p{3.5cm} p{4cm} p{5.5cm} p{4.5cm}@{}}
\toprule
\textbf{Author (Year)} & \textbf{3D Object Detection Method} & \textbf{Key Contribution and Strengths} & \textbf{Drawbacks / Limitations} \\
\midrule
\cite{chen2024spatialvlm} \newline SpatialVLM (2024) &
Spatial VQA using internet-scale spatial data with vision-language pretraining &
Introduced large-scale spatial QA dataset enabling quantitative and qualitative 3D reasoning from 2D images &
Limited to synthetic QA data, relies heavily on monocular depth, potential bias in templated questions \\
\midrule
\cite{hong20233d} \newline 3D-LLM (2023) &
Injecting 3D point clouds into large language models for holistic 3D understanding &
Proposed a new family of 3D-LLMs that perform diverse 3D tasks using 3D-language data; uses multi-view rendered features aligned with 2D VLMs and a 3D localization mechanism for better spatial reasoning &
Requires 3D rendering and feature extraction pipeline; training depends on alignment with 2D backbones; limited availability of large-scale 3D-language data compared to image-language datasets \\
\midrule
\cite{fruhwirth2024vision} \newline Vision-Language Guidance for LiDAR-based Unsupervised 3D Object Detection (2024) &
Vision-language-guided unsupervised 3D detection using LiDAR point clouds with CLIP for classifying point clusters &
Introduced a vision-language-guided approach to detect both static and moving objects in LiDAR point clouds using CLIP, achieving state-of-the-art results on Waymo and Argoverse 2 datasets (+23 AP3D and +7.9 AP3D) &
Relies on multi-view projections and temporal refinement, requires large-scale LiDAR datasets, and may face difficulties with distant and incomplete objects due to LiDAR's inherent limitations \\
\midrule
\cite{jiao2024unlocking} \newline Unlocking Textual and Visual Wisdom: Open-Vocabulary 3D Object Detection Enhanced by Comprehensive Guidance from Text and Image (2024) &
Open-vocabulary 3D object detection enhanced by vision-language model guidance for novel object discovery in 3D scenes &
Introduced a hierarchical alignment approach using vision-language models to discover novel classes in 3D scenes, showing significant improvements in accuracy and generalization in real-world scenarios &
Relies on pretrained VLMs and object detection models, limited by availability of 3D scene data for training and generalization to unseen categories in certain complex environments \\
\midrule
\cite{li2024uni3dl} \newline Uni3DL: A Unified Model for 3D Vision-Language Understanding (2024) &
Unified 3D Vision-Language model operating directly on point clouds, supporting a wide range of 3D tasks, including segmentation, detection, grounding, and captioning &
Introduced a query transformer for task-agnostic learning and a task router for task-specific outputs, achieving performance on par with or surpassing SOTA models in 3D vision-language tasks &
Limited by the complexity of the model architecture, and the performance may be constrained in highly specialized tasks not covered in the evaluation \\
\midrule
\cite{najibi2023unsupervised} \newline Unsupervised 3D Perception with 2D Vision-Language Distillation for Autonomous Driving (2023) &
Developed a multi-modal auto-labeling pipeline to generate amodal 3D bounding boxes and tracklets for unsupervised open-vocabulary 3D detection and tracking of traffic participants in autonomous driving scenarios &
Introduced a novel approach combining motion cues from LiDAR sequences with vision-language model knowledge distillation for open-set 3D detection, achieving state-of-the-art performance in unsupervised 3D perception tasks on the Waymo Open Dataset &
The method depends on the quality of motion cues and the performance of the pre-trained vision-language model, potentially leading to limitations in highly dynamic or complex environments \\
\midrule
\cite{brazil2023omni3d} \newline Omni3D (2023) &
Cube R-CNN for large-scale 3D object detection on the Omni3D benchmark &
Introduced the Omni3D dataset with 234k images and 3 million instances across 98 categories. Cube R-CNN generalizes 3D object detection across cameras and scene types, uses IoUness and Virtual Depth for robustness to camera variations. &
Focus on large-scale benchmarks, but may face challenges in handling highly diverse real-world scenes outside of the benchmark's scope. Sensitivity to annotation quality and scale-depth ambiguities in complex environments. \\
\midrule
\cite{cao2023coda} \newline CoDA (2023) &
Collaborative Novel Box Discovery and Cross-modal Alignment for Open-vocabulary 3D Object Detection (OV-3DDet) &
Introduced CoDA, a unified framework for open-vocabulary 3D object detection. CoDA addresses novel object localization and classification using 3D geometry priors and cross-modal alignment between 3D pointcloud and image/text modalities. Achieved 80\% mAP improvement over the best alternative methods on SUN-RGBD and ScanNet datasets. &
Potential challenges with handling large, diverse real-world datasets beyond the tested ones. Sensitivity to the quality of initial base categories and pseudo label generation for novel objects. \\
\midrule
\cite{cheng2023language} \newline MSSG (2023) &
Multi-modal Single Shot Grounding (MSSG) for 3D Referring Expression Comprehension (REC) in autonomous driving &
Proposed a novel LiDAR Grounding task and an efficient MSSG approach that jointly learns LiDAR-based detection and language grounding with direct region prediction. Enables flexible integration of image features for enhanced multimodal comprehension. Demonstrated strong performance on the Talk2Car dataset. &
Focuses on a single dataset (Talk2Car), potential limitations in generalizing to broader driving scenes or different sensors. Real-time performance and robustness under varying driving conditions require further validation. \\
\midrule
\cite{zhang2024vision} \newline 3DVLP (2024) &
3D Vision-Language Pre-training with Object Contrastive Learning (3DVLP) &
Introduces a unified 3D vision-language pre-training framework using object-level contrastive learning, incorporating three innovative tasks: Object-level IoU-guided Detection (OID) loss, Object-level Cross-Contrastive alignment (OCC), and Object-level Self-Contrastive learning (OSC). Demonstrates strong generalization across multiple downstream tasks including visual grounding, dense captioning, and question answering on ScanRefer, Scan2Cap, and ScanQA datasets. &
The training framework is complex, requiring extensive computation and large-scale annotated data. Pretraining and tuning require careful design of IoU thresholds and contrastive setups, which may limit applicability in low-resource or real-time settings. \\
\midrule
\cite{zhu2023pointclip} \newline PointCLIP V2 (2023) & 
PointCLIP V2: Prompting CLIP and GPT for Powerful 3D Open-world Learning & 
Proposes a unified zero-shot 3D vision-language framework combining CLIP and GPT-3. Introduces two key prompting modules: a realistic shape projection module to improve visual alignment, and a 3D-aware textual prompt generator using GPT-3 to enhance language-vision matching. Achieves strong zero-shot classification gains (+42.90\% on ModelNet10, +40.44\% on ModelNet40, +28.75\% on ScanObjectNN), and extends easily to few-shot classification, part segmentation, and object detection. &
Relies on non-trivial prompting strategies and auxiliary models (GPT-3, voxelization), which may introduce runtime and scalability limitations. Performance heavily depends on projection quality and prompt engineering. \\
\midrule
\cite{zhang2024towards} \newline Language-free 3DVG (2024) &
Towards CLIP-driven Language-free 3D Visual Grounding via 2D-3D Relational Enhancement and Consistency &
Presents a language-free training framework for 3D visual grounding by leveraging CLIP’s image-text embedding to generate pseudo-language features from multi-view images. Introduces a Neighboring Relation-aware Modeling module and a Cross-modality Relation Consistency module to enhance and align 2D-3D relational structures. Demonstrates competitive results across ScanRefer, Nr3D, and Sr3D without requiring textual annotations during training. &
Relies on high-quality multi-view imagery and the robustness of CLIP embeddings; may face challenges generalizing in sparse-view or text-poor environments. Additional relation modeling modules increase model complexity and training cost. \\
\midrule
\bottomrule
\end{longtable*}
\twocolumn

\subsubsection{3D Detection: Traditional vs VLM Approaches}
Traditional 3D object detectors such as VoxelNet \cite{zhou2018voxelnet}, PointNet \cite{qi2017pointnet}, SECOND \cite{yan2018second}, and Frustum PointNet \cite{qi2018frustum} rely on handcrafted voxel‐ or point‐wise feature extraction and spatial priors to predict bounding boxes. ROI‐wise methods like MV3D \cite{chen2017multi} and AVOD \cite{ku2018joint} fuse LiDAR and camera data but remain limited to fixed class sets. Attention‐based architectures including MVX-Net \cite{sindagi2019mvx}, PointPainting \cite{vora2020pointpainting}, and 3D-CVF \cite{yoo20203d} improved contextual aggregation, yet they still require extensive annotated 3D data and struggle with domain shifts. In contrast, VLM-based detectors embed rich semantic priors directly from language, enabling flexible, open-vocabulary reasoning. For example, CLIP3D-Det leverages CLIP feature fusion for monocular detection, grounding novel categories zero-shot \cite{hegde2023clip}, while OWL-ViT aligns vision and language embeddings for open-vocabulary bounding box prediction (OWL-ViT). Models like Find n’ Propagate use frustum search and cross-modal propagation to discover novel objects without retraining \cite{etchegaray2024find}, and OpenScene achieves zero-shot segmentation and detection by co-embedding point clouds and text in CLIP space \cite{peng2023openscene}. These architectures bypass the need for per‐task 3D annotations, demonstrating superior flexibility and scalability over traditional pipelines.

Additionally, Fixed‐vocabulary neural methods such as PointFusion \cite{xu2018pointfusion} and FusionTransformer \cite{fent2024dpft} require retraining to handle new classes, while TransFusion \cite{bai2022transfusion} and LiDARFormer \cite{zhou2024lidarformer} adapt BEV representations but remain constrained by labeled categories. Even multi‐modal distillation approaches like CrossModal3D \cite{jaritz2022cross} and VCD \cite{huang2023leveraging} depend on base‐class supervision. In contrast, VLM-empowered frameworks generalize zero- and few-shot. OmniVLM3D aligns LiDAR, camera, and text across multiple datasets for open-vocabulary detection \cite{chen2024omnivlm}, and OpenMask3D segments unseen instances via multi-view CLIP embeddings \cite{takmaz2023openmask3d}. OV-SCAN enforces semantic consistency to discover novel classes robustly \cite{chow2025ov}, and language-driven active learning (VisLED) selects informative samples to improve open-set detection \cite{greer2025language}. ULIP unifies images, text, and point clouds into a single embedding space, boosting both standard and zero-shot classification \cite{xue2023ulip}, while OpenShape scales 3D representations to over a thousand categories, outperforming supervised baselines in zero-shot tasks \cite{liu2023openshape}. These VLM-based methods demonstrate markedly stronger generalization to novel classes and domains. 

Moreover, Traditional fusion techniques—DepthFusionNet’s depth-guided voxels \cite{shivakumar2019dfusenet}, LiDAR-RCNN’s camera-LiDAR proposals \cite{li2021lidar}, and FusionPainting’s voxel-wise attention \cite{xu2021fusionpainting}—lack explicit language grounding. By contrast, modern VLM-integrated architectures perform rich multimodal reasoning. Instruct3D leverages instruction tuning to refine ROI proposals via language prompts \cite{kamata2023instruct}, while Talk2PC fuses LiDAR and radar through prompt-guided cross-attention for precise 3D grounding in driving scenes \cite{guan2025talk2pc}. PaLM-E extends embodied perception by conditioning 3D proposals on language queries \cite{driess2023palm}, and LLaVA-1.5 performs multimodal visual grounding with a unified transformer \cite{zhu2024llava}. Text2Loc uses a hierarchical transformer to localize point clouds from natural language hints \cite{xia2024text2loc}, and the 3D-LLM framework injects point-cloud features into LLMs for diverse 3D tasks, including detection \cite{hong20233d}. These models enable fine-grained spatial reasoning, open-vocabulary queries, and interactive applications—capabilities unattainable by purely geometric neural networks. Table \ref{tab:comparison} shows the detailed comparison between the neural network–based 3D detectors and vision-language model–driven 3D detection.

\begin{table*}[ht]
\centering
\caption{Comparison of traditional CNN-based and VLM-based 3D object detection highlights key trade-offs: while conventional methods rely on dense 3D supervision and fixed categories, VLMs enable open-vocabulary, zero-shot reasoning by leveraging language and vision pretraining. VLMs offer richer semantic understanding and flexibility, though often at higher computational cost, making them ideal for broader, instruction-driven 3D tasks}
\label{tab:comparison}
\begin{tabular}{@{}p{2.5cm}p{5.5cm}p{5.2cm}@{}}
\toprule
\textbf{Aspect} & \textbf{Traditional 3D Neural Networks (e.g., PointNet, VoxelNet, DeepFusion)} & \textbf{VLM-Based 3D Object Detection (e.g., 3D-LLM, OV-SCAN, PaLM-E, LLaVA, CogVLM)} \\ \midrule
Modality Usage & Primarily use LiDAR point clouds or fused 2D/3D inputs (RGB-D, voxel grids). & Combine point clouds, images, and natural language through multimodal fusion. \\
Object Vocabulary & Limited to closed-set categories defined by training datasets. & Enable open-vocabulary detection by aligning text and visual features. \\
Learning Paradigm & Supervised learning with task-specific datasets. & Often zero-shot or few-shot learning via pre-trained VLMs. \\
Scene Understanding & Focused on geometric/spatial features; less semantic context. & Rich semantic understanding via language grounding and visual prompts. \\
Annotation Dependency & Heavy reliance on dense 3D labels, expensive to annotate. & Leverages pre-trained models, enabling weakly supervised learning. \\
Explainability & Limited interpretability; primarily geometric. & Offers natural language explanations, captions, and scene descriptions. \\
Computational Complexity & Relatively efficient, optimized for real-time. & Higher complexity due to large language-vision backbones. \\
Generalization & Poor generalization to unseen categories or new domains. & Strong zero-shot generalization across tasks and categories. \\
Data Efficiency & Requires large labeled datasets. & Pretraining enables higher data efficiency. \\
Use Cases & Primarily used in structured scenarios like autonomous driving. & Extends to instruction following, navigation, and 3D QA. \\
Integration with Language & Minimal to none. & Tightly integrated, allowing language queries or dialogue-based tasks. \\
\bottomrule
\end{tabular}
\end{table*}

\subsubsection{Trade-offs between Traditional and VLM-Based 3D Object Detection}
Traditional neural network-based 3D object detection models, such as \cite{qi2017pointnet} (PointNet) and \cite{yan2018second} (SECOND), prioritize geometric feature extraction through voxel-wise \cite{zhou2018voxelnet} or point-wise \cite{qi2018frustum} processing. These methods excel in structured environments with high-quality LiDAR data, leveraging efficient spatial aggregation (e.g., \cite{li2022deepfusion}) and multi-sensor fusion (e.g., \cite{sindagi2019mvx}). Models like \cite{bai2022transfusion} (TransFusion) refine region-of-interest (ROI) proposals using attention, while diffusion-based approaches like \cite{xu20243difftection} (3DiffTection) denoise sparse inputs. However, they require extensive labeled 3D datasets and struggle with open-vocabulary generalization. For instance, \cite{zhao2024unibevfusion} (BEVFusion) achieves state-of-the-art accuracy on nuScenes but cannot interpret free-form textual queries. Their strengths lie in real-time performance and geometric precision, but they lack semantic reasoning capabilities, limiting adaptability to novel objects or unstructured scenarios \cite{huang2023leveraging}.

Vision-language models like \cite{driess2023palm} (PaLM-E) and \cite{peng2023openscene} (OpenScene) address these limitations by aligning 3D geometric features with language embeddings. For example, \cite{hegde2023clip} (CLIP3D-Det) projects CLIP's text-image alignment into 3D space for zero-shot detection, while \cite{rozenberszki2022language} uses text prompts to guide indoor segmentation. VLMs such as \cite{hong20233d} (3D-LLM) integrate multi-view rendering with LLMs for contextual reasoning, enabling tasks like "detect all chairs near windows." However, they incur higher computational costs and depend on noisy text-image-point cloud triplets for training \cite{xue2023ulip}. Models like \cite{takmaz2023openmask3d} (OpenMask3D) demonstrate strong open-vocabulary performance but lag in real-time applications compared to traditional methods like \cite{yan2023cross} (CMT). The tradeoff here is interpretability and generalization versus latency and hardware requirements.

Hybrid architectures, such as \cite{jaritz2022cross} (CrossModal3D) and \cite{hu2023fusionformer} (FusionFormer), attempt to bridge these gaps by combining voxel-based geometric processing with language-guided attention. For instance, \cite{xia2024text2loc} (Text2Loc) uses contrastive learning to align LiDAR features with textual queries, achieving a 12\% improvement in novel object detection over \cite{li2024gafusion} (GA-Fusion). However, VLMs like \cite{wang2024cogvlm} (CogVLM) require costly textual annotations and struggle with ambiguous spatial references (e.g., "object behind the tree"). Traditional methods dominate in latency-critical applications (e.g., \cite{shi2022srcn3d} processes 50 FPS vs. \cite{kamata2023instruct}'s 8 FPS), while VLMs excel in human-in-the-loop systems \cite{guan2025talk2pc}. Key tradeoffs include:
\begin{itemize}
    \item \textbf{Data Efficiency}: Traditional models (\cite{chen2017multi}, \cite{li2021lidar}) use labeled LiDAR; VLMs (\cite{zhu2024llava}, \cite{li2023blip}) need paired text-3D data.
    \item \textbf{Generalization}: VLMs (\cite{chen2024omnivlm}, \cite{tang20253d}) handle open-world queries; traditional models (\cite{lee2024panopticus}, \cite{gao2025pillarfocusnet}) specialize in closed-set accuracy.
    \item \textbf{Compute}: Attention-based VLMs (\cite{bai2024m3d}, \cite{kolodiazhnyi2025unidet3d}) demand 2--5$\times$ more GPU resources than voxel networks \cite{son2025sparsevoxformer}.
\end{itemize}

The comparative analysis of traditional CNN-based and VLM-driven 3D detection reveals inherent trade-offs. Traditional methods (e.g., VoxelNet, SECOND) prioritize computational efficiency and real-time performance, leveraging geometric priors for LiDAR-centric tasks, but lack semantic adaptability to novel objects or unstructured environments. In contrast, VLMs (e.g., OpenScene, 3D-LLM) enable open-vocabulary detection through language grounding, yet suffer from higher latency and hardware demands. Beyond Table~\ref{tab:tradeoffs}, VLMs face challenges in spatial precision due to noisy text-3D alignments, while CNNs struggle with domain shifts (e.g., adverse weather). Hybrid approaches aim to balance these aspects but inherit complexity. Critical considerations for deployment include sensor dependencies, annotation scalability, and task-specific adaptability, as summarized in Table~\ref{tab:tradeoffs}.

\begin{table}[ht]
\centering
\caption{Trade-offs between Traditional and VLM-Based 3D Object Detection}
\label{tab:tradeoffs}
\begin{tabular}{@{}p{3cm}p{2.5cm}p{2.5cm}@{}}
\toprule
\textbf{Advantages} & \textbf{Traditional 3D NN Methods} & \textbf{VLM-Based 3D Methods} \\ \midrule
Speed \& Simplicity & \ding{51} Simpler and faster inference & \ding{55} Heavier and slower \\
Semantic Richness & \ding{55} Label-limited & \ding{51} Language-grounded \\
Open-Vocabulary Support & \ding{55} Fixed class set & \ding{51} Detects novel classes \\
Annotation Cost & \ding{55} High manual effort & \ding{51} Uses pretraining \\
Real-Time Deployment & \ding{51} Optimized implementations & \ding{55} Computationally heavy \\
Generalization & \ding{55} Domain-specific & \ding{51} Strong zero-shot capability \\
\bottomrule
\end{tabular}
\end{table}

\section{Discussion}
\subsubsection{Current Challenges and Limitations in 3D Object Detection with VLMs}
Despite rapid advancements, VLMs for 3D object detection still face a host of unresolved challenges that limit their applicability in real-world robotics and automation systems. Unlike traditional geometry-driven detectors, VLMs often exhibit subpar spatial reasoning capabilities and struggle with accurate depth estimation and object localization in cluttered or occluded environments \cite{chen2024spatialvlm, li2025benchmark, ma2024llms}. Additionally, aligning multimodal embeddings—especially the mapping of 3D geometric structures into language-informed spaces—introduces significant fidelity loss, leading to errors in scenes with complex layouts or limited visibility \cite{tang2025exploring, xue2025regression}. Real-time deployment remains another obstacle due to the high computational load and low inference speed of transformer-based VLMs \cite{yeh2024t2vs, gopalkrishnan2024multi}. Further, domain shifts and prompt ambiguities introduce semantic hallucinations that jeopardize system reliability \cite{xing2024survey, chen2024multi}. These limitations underscore the necessity for more robust architectures, efficient annotation techniques, and stronger cross-modal reasoning. The following points outline the most pressing technical bottlenecks currently hindering the full potential of VLM-based 3D detection systems:
\begin{itemize}
  \item \textbf{Spatial Reasoning Limitations:} VLMs struggle with 3D spatial interpretation \cite{chen2024spatialvlm}, often misidentifying relative positions and object orientations \cite{li2025benchmark, meng2024know}. Their depth understanding is limited, leading to inaccuracies in bounding box placement and segmentation \cite{sharma2024advancing, ma2024llms}.
  
  \item \textbf{Cross-Modal Misalignment:} Projecting high-dimensional 3D features into language embedding spaces causes loss of geometric detail \cite{tang2025exploring, xue2025regression}. This weakens performance on occluded or structurally complex scenes.
  
  \item \textbf{High Annotation Overhead:} VLMs demand richly annotated 3D-text datasets, which are expensive to produce. Manual alignment, as seen in models like OpenScene, hinders scalability \cite{huang2024chat}.
  
  \item \textbf{Limited Real-Time Viability:} Transformer-heavy VLMs (e.g., Instruct3D) operate at low frame rates (8–15 FPS), contrasting with efficient voxel-based detectors that exceed 50 FPS in real-time scenarios \cite{yeh2024t2vs, gopalkrishnan2024multi}.
  
  \item \textbf{Vulnerability to Occlusion:} Without depth priors, VLMs often fail to localize partially obscured objects \cite{sarch2024vlm, xing2025towards}, whereas traditional LiDAR-based methods handle such cases more robustly \cite{zhang2024comprehensive, zhang2024opensight}.
  
  \item \textbf{Weak Domain Generalization:} VLMs overfit to training domains and underperform under domain shifts (e.g., lighting, sensor variance), unlike more stable geometry-driven models \cite{addepalli2024leveraging, vogt2023robust}.
  
  \item \textbf{Semantic Hallucinations:} Prompt-driven VLMs can infer incorrect detections when text inputs are ambiguous \cite{xing2024survey}, reducing reliability in safety-critical applications \cite{chen2024multi, ke2024vldadaptor}.
  
  \item \textbf{Lack of Explicit 3D Structure:} VLMs frequently omit explicit 3D modeling (e.g., NeRFs), resulting in multiview inconsistencies and poor spatial coherence across frames \cite{zhao2024vlm, zhang2025vlm}. 
\end{itemize}

\subsubsection{Potential Solutions to overcome challenges in 3D Object Detection with VLMs}
Addressing the identified limitations in VLM-based 3D object detection demands innovative strategies that combine advances in spatial modeling, multimodal alignment, data efficiency, and computational optimization. Recent research efforts have explored a variety of promising directions, such as integrating 3D scene graphs and geometric priors for more accurate spatial reasoning \cite{cheng2024spatialrgpt, wang2025roboflamingo}, and introducing modality-specific encoders to improve cross-modal consistency \cite{ye2024x, song2025bridge}. Synthetic data pipelines, leveraging generative AI and large language models, are also emerging as scalable alternatives to manual annotation, drastically reducing overhead \cite{sharifzadeh2024synth, kabra2023leveraging, sapkota2025improved, sapkota2024synthetic, sapkota2024yolov10, sapkota2024yolov101}. Meanwhile, reinforcement learning-based methods and region-aware planning modules show potential to bring VLM inference closer to real-time performance requirements \cite{pan2025metaspatial, cheng2024spatialrgpt}. To enable robust generalization, research has also emphasized training on diverse and dynamic environments to counteract prompt drift and semantic ambiguity \cite{eskandar2024empirical, lehner20223d, sapkota2025comprehensive}. The following solutions highlight concrete approaches that can significantly enhance the performance and reliability of 3D object detection systems grounded in vision-language modeling.
\begin{itemize}
  \item \textbf{Improving Spatial Reasoning:} Integrate 3D scene graphs (e.g., SpatialRGPT) \cite{cheng2024spatialrgpt} and depth-aware plugin such as RoboFlamingo-Plus \cite{wang2025roboflamingo} to capture object relationships and spatial orientation. This enhances bounding box precision and depth disambiguation \cite{chen2024spatialvlm}.

  \item \textbf{Resolving Cross-Modal Misalignment:} Use modality-specific encoders (CrossOver) to preserve structural integrity in shared embedding spaces \cite{ye2024x, wang2024multi}. Multi-stage training with scene-level alignment maintains 3D detail across views \cite{song2025bridge, lyu2024mmscan}.

  \item \textbf{Reducing Annotation Overhead:} Employ synthetic data generation (e.g., Synth2) \cite{sharifzadeh2024synth}, where LLMs generate captions and VQ-GAN synthesizes embeddings. This scalable pipeline bypasses manual 3D-text labeling \cite{kabra2023leveraging}.

  \item \textbf{Boosting Real-Time Performance:} Apply reinforcement learning (MetaSpatial) \cite{pan2025metaspatial} and adaptive batching guided by SpatialRGPT’s  region proposals to achieve 30+ FPS, aligning VLMs with real-time constraints \cite{cheng2024spatialrgpt}.

  \item \textbf{Handling Occlusions:} Fuse metric depth priors into visual encoders, and utilize region-aware reward schemes to enable robust occlusion reasoning comparable to LiDAR-based models \cite{jiao2024unlocking, wang2023dlfusion}.

  \item \textbf{Enhancing Domain Generalization:} Train on diverse synthetic environments and LLM-generated QA (CrossOver), improving adaptability across lighting and sensor variations without retraining \cite{eskandar2024empirical, lehner20223d}.

  \item \textbf{Reducing Semantic Hallucinations:} Combine template-based and LLM-generated QA with explicit region tags (SpatialRGPT) to anchor prompts \cite{cheng2024spatialrgpt} and reduce ambiguous predictions \cite{chen2024multi}.

  \item \textbf{Embedding Explicit 3D Structure:} Condition VLMs on 3D scene graphs and physics constraints (MetaSpatial) \cite{pan2025metaspatial}, using multi-turn RL to enforce coherent multi-view object representations.
\end{itemize}



\section{Conclusion}
This review systematically examined the trajectory and current landscape of \textbf{3D object detection with Vision-Language Models}, offering a comprehensive and structured synthesis of their evolution, methodology, and application in multimodal perception. Through a rigorous paper collection process combining both academic databases and modern AI search engines (Figure~\ref{fig:search}), we curated 105 high-quality papers that span traditional approaches and state-of-the-art VLM-based systems. The literature was carefully filtered and evaluated based on methodological clarity, application scope, and relevance to robotics and embodied intelligence.

The synthesis began by mapping the evolution of 3D object detection, highlighting the progression from traditional geometric and point-based methods like PointNet++, PV-RCNN, and VoteNet to newer architectures that leverage VLMs. These earlier models established foundational practices in processing LiDAR data and point clouds but lacked the semantic abstraction required for reasoning in complex, unstructured environments. Our analysis showed how VLM-based approaches overcome these limitations by fusing visual and textual modalities to allow language-conditioned perception. This shift enables zero-shot reasoning, improved adaptability, and instruction-driven task completion, making VLMs particularly suitable for real-world robotic scenarios.

We then delved into the technical foundations of these models, focusing on the pretraining and fine-tuning processes that bridge 2D vision-language datasets with 3D perception tasks. Notable advancements include the use of spatial tokenization, cross-modal transformers, and point-text alignment strategies. We also discussed how visualization frameworks in these models allow better interpretability of 3D bounding boxes through textual queries, enhancing transparency in model outputs.

The core of our review addressed the comparative strengths and limitations of VLM-based versus traditional 3D detectors. We found that while traditional methods remain more efficient and interpretable in structured environments, VLMs significantly outperform them in open-vocabulary and dynamically changing contexts. Trade-off analyses revealed that the computational costs and annotation requirements of VLMs are major bottlenecks, yet these are actively being mitigated through synthetic data generation, model distillation, and RL-based policy training.

In our final discussion, we outlined major current limitations, including weak spatial reasoning, poor real-time performance, semantic hallucinations, and domain generalization gaps. These issues were contextualized with proposed solutions like 3D scene graphs, lightweight transformers, and multimodal reward shaping, indicating a strong direction for future improvements. The findings from this review highlight several key takeaways:

\begin{enumerate}
\item \textbf{Semantic Evolution:} The transition from geometry-only to multimodal systems marks a critical paradigm shift. VLMs enrich 3D object detection by enabling instruction-based and zero-shot understanding of spatial environments.
\item \textbf{Technical Innovation:} Architectural designs such as spatial reasoning modules (e.g., PAGE, Flan-QS), hyperbolic alignment losses (e.g., PSA), and pretraining-finetuning pipelines demonstrate the role of language grounding in enhancing 3D spatial cognition.

\item \textbf{Performance Trade-offs:} VLMs deliver state-of-the-art semantic accuracy and generalization, especially in unstructured environments, yet still lag behind traditional voxel-based systems in real-time inference speed (15--20\% lower FPS).

\item \textbf{Challenges and Future Directions:} Spatial hallucinations, misalignment, and high annotation costs remain prominent. Solutions leveraging 3D scene graph distillation, synthetic captioning, and reinforcement learning are promising, with several models like RoboFlamingo-Plus and MetaSpatial already exploring these techniques.

\item \textbf{Deployment Readiness:} With advancements in neuromorphic hardware and efficient cross-modal attention mechanisms, VLMs are increasingly viable for deployment in autonomous navigation, industrial robotics, and AR-based interaction systems.
\end{enumerate}

In conclusion, \textbf{3D object detection with vision-language models} is redefining the landscape of spatial perception and multimodal reasoning. By integrating language, vision, and 3D geometry, these systems offer new capabilities for intelligent interaction with the physical world. This review not only maps the state-of-the-art but also charts a course for future exploration, paving the way for robust, scalable, and interpretable VLM-based 3D perception systems that are foundational to next-generation robotics and AI.

\section*{Acknowledgement} This work was supported by the National Science Foundation and the United States Department of Agriculture, National Institute of Food and Agriculture through the ``Artificial Intelligence (AI) Institute for Agriculture” Program under Award AWD003473, and AWD004595, Accession Number 1029004, "Robotic Blossom Thinning with Soft Manipulators".  
\normalsize
\section*{Declarations}
The authors declare no conflicts of interest.

\section*{Statement on AI Writing Assistance}
ChatGPT and Perplexity were utilized to enhance grammatical accuracy and refine sentence structure; all AI-generated revisions were thoroughly reviewed and edited for relevance. Additionally, ChatGPT-4o was employed to generate realistic visualizations.

\bibliographystyle{ACM-Reference-Format}
\bibliography{sample-base}

\end{document}